\documentclass[10pt,lettersize,journal]{IEEEtran}
\usepackage[T1]{fontenc}
\usepackage[utf8]{inputenc}
\usepackage{amsmath,amssymb,graphicx,booktabs,url}
\usepackage{cite}
\usepackage{pgfplots}
\pgfplotsset{compat=1.17}
\definecolor{plotcolor0}{HTML}{0072B2}
\definecolor{plotcolor1}{HTML}{E69F00}
\definecolor{plotcolor2}{HTML}{009E73}
\definecolor{plotcolor3}{HTML}{D55E00}
\definecolor{plotcolor4}{HTML}{CC79A7}
\usepackage[hidelinks]{hyperref}
\newsavebox{\uploadedtablebox}
\title{Regional Complementarity and Learned Weighting for Occluded Face Verification}
\author{Sangiya Pararajasingham\thanks{Independent Researcher. ORCID: 0009-0005-3407-2154.}}
\begin{document}
\maketitle
\begin{abstract}
We explore the complementary nature of different facial regions in occluded face verification. Multiple DINOv2 fusion approaches, including no fusion, fixed fusion, and learned fusion, were employed. Five model configurations were trained; frozen encoders with trained projection layers, full face LoRA, periocular LoRA, and dual branch models with equal fusion weights/learned fusion. All models were trained with three seeds on the identity-disjoint ROF training split (4405 of 6171 manifest images) and evaluated using frozen ROF, MFR2, and RMFRD verification protocols. The best performance for masked data was found in the periocular LoRA model (0.9221 \ensuremath{\pm} 0.0046). For sunglasses, the best performance was with the fixed fusion approach (0.8950 \ensuremath{\pm} 0.0021). Neutral data had the best performance in the learned fusion model (0.9458 \ensuremath{\pm} 0.0026). With learned fusion, 0.9548 \ensuremath{\pm} 0.0022 performance was achieved in 802 MFR2 pair comparisons after a known training-name overlap was excluded. Fixed fusion achieved best performance (0.7240 \ensuremath{\pm} 0.0099) from 21798 RMFRD pair comparisons with exact content duplicates removed. When ensuring that the branches in the model were consistently trained, a varying weighted fusion does not guarantee performance gains over a constant weighted fusion. Validation-calibrated thresholds reveal shifts in test false acceptance rates that retrospective AUC summaries do not establish. Separate aligned-SFace and shared-crop SFace evaluations show sensitivity to preprocessing; their AUCs do not establish architectural superiority. Evidence of complementary data in different conditions is provided. We show that using learned fusion gates with high AUCs alone does not guarantee~\mbox{reliable performance} in \mbox{occluded verification}.\par
\end{abstract}
\section{Introduction}
Face verification determines if two images represent the same individual. In cases where there are occlusions, in either probe or enrolment, the appearance of the face will change. Masked faces result in reduced information from the bottom half of the face. Sunglasses result in reduced information from the eyes and surrounding skin. Therefore, information can be lost between different types of occlusion. Controlling false acceptance rates is crucial in such scenarios.\par
While periocular regions allow for some top of face information in the masked scenarios, there may be useful features within the neutral images. The periocular regions lose information when sunglasses obscure the eyes. Therefore, using both full face and periocular representations seems reasonable. However, having multiple branches in a network does not mean that learning how to best fuse them will lead to improved verification performance.\par
Multiple different representations were tested. Different models include (1) Frozen full face representation, with learned projection; (2) Full face with LoRA; (3) Periocular with LoRA; (4) Two branch model with equal weighting; and (5) Learned fusion. All 5 configurations were independently trained and used the same frozen verification pairs for evaluation. Three seeds were used in training these models. The raw scores and signatures can be followed to see the results. This was done to understand what factors (adaptation, region choice, and fusion) improve performance.\par
As different weights are applied in the fixed and gated fusion models, different features are being learnt in each branch. By applying different weights to the gated models, but keeping the same branches, it was found that equal weighting in the gated branches performs better than learned weighting in masked ROF pairs.\par
To address issues in historical data, known exact-content conflicts and both copies of one confirmed cross-split near-duplicate photograph were removed. After retraining on the cleaned ROF manifest, external evaluation excluded the known exact-name overlap between MFR2 and ROF training and the recorded cross-label exact-content conflicts in RMFRD. This means that certain issues have been addressed in the data. However, we cannot be confident that all semantically meaningful labels are valid, nor that data is independent from backbone pre-training.\par
Three key findings were observed. First, comparisons between configurations across different seeds help distinguish effects of adaptation, region selection, and fusion in fixed pairs. Second, changing the weights within the same branches allows us to understand if improved performance in learned representations is due to the inference weighting rule. Third, thresholding based on validation data, sensitivity to identity removal and using shared-crop face-recognition models as controls indicate that limitations in performance exist beyond what would be obvious from comparing AUCs. Overall, improvements in replicability have been made. However, it is not novel to perform adaptation in DINOv2 models \cite{dinov2} using LoRA \cite{lora}, nor to exploit complementary information between full face and periocular representations.\par
\section{Related Work}
\subsection{Face verification}
Embeddings are learnt such that images of the same individual cluster together. Distinct individuals should be separated. Cosine similarity between embeddings is commonly used as it is easy to implement, invariant to scaling (once normalised) and is compatible with ROC analysis. Improvements in clean face verification benchmarks have been made by employing margin based losses. SphereFace uses a multiplicative angular margin \cite{sphereface}, CosFace uses an additive margin in cosine similarity \cite{cosface} and ArcFace uses additive margin in the hypersphere \cite{arcface}. AdaFace allows for adaptive margins dependent on image quality \cite{adaface}. MagFace also incorporates adaptive margins, depending on embedding magnitude to reflect quality \cite{magface}. During training, only known identities are optimised. In ROF verification, unknown identities are verified. In external protocols, limits on identity overlap are imposed.\par
\subsection{Masked and occluded face recognition}
Since the beginning of the pandemic, masked face recognition has become an area of specific interest. Masked face simulation (MaskTheFace) and masked face datasets (MFR2) were introduced \cite{mfr2}. MAFA/LLE-CNNs were proposed by Ge et al. to detect masked faces \cite{llecnn}. Wang et al. created a series of masked face datasets (MFDD, RMFRD, SMFRD) \cite{rmfrd}. A review of occlusion robust face recognition approaches was conducted in 2021 \cite{surveyocc}. Recently, dynamic masking of corrupted features \cite{from2022}, pairwise differential Siamese mask learning \cite{masklearn} and self restrained triplet loss to unmask face embeddings \cite{umr} have been proposed. A multi task contrastive learning approach was proposed by Neto et al. called FocusFace \cite{focusface}. Neto et al. also analyse generalisability of masked face models to broader occlusions \cite{beyondmasks}. Strong results have been found using real masked face datasets. Strong results were achieved in IJCB competition in MFR 2021 \cite{mfr2021}, and in the InsightFace track \cite{insightfacemfr}. Training models on synthetic masked faces and testing on real masks has been investigated by Hsu et al. \cite{hsu2022}. Real upper (sunglasses) and lower (mask) occlusions, with specific evaluation protocols, exist in the ROF dataset \cite{rof}. Recently, direct complementary information between face and periocular representations has also been considered. Attention and prompting are used to learn aligned face and periocular representations in Flexible Biometrics Recognition \cite{fbr2024}. This is especially applicable to this work as complementary information between these two regions was considered.\par
\subsection{Periocular recognition}
Periocular biometric information can be extracted. Both visible spectrum periocular comparison (including eyebrow information) and merging face and periocular information were considered by Park, Jillela, Ross, and Jain \cite{periocular}. Recently, consistent knowledge distillation has been applied to allow periocular embeddings to be trained from full face data. This means useful information can be passed to periocular embeddings even if only periocular data is available at inference time \cite{ckd2024}. Periocular recognition is therefore a suitable option to handle lower occlusions. However, sunglass occlusions impact the periocular region.\par
Recently, Alonso-Fernandez et al. trained periocular networks using ocular cropped regions from large face datasets \cite{ocularcrop2025}. This highlights the importance of considering the size of training data, how cropping is performed, and acquisition protocol between full face and periocular models. However, results cannot be directly compared between this method of verification and fixed pair verification in ROF. Prasad et al. used LoRA adapted DINOv2 models with quality aware fusion of enrolment templates to enable cross spectral periocular recognition \cite{perilora2026}. In contrast, in this work, the weighting between full face and periocular branches in visible spectrum occlusions is explored. Constant auxiliary inputs and negative gate controls are employed. While it is not novel to apply DINOv2 with LoRA in periocular recognition, nor the quality aware fusion approach, this work is not the first to consider periocular recognition with DINOv2 and LoRA.\par
\subsection{Pretrained visual representations}
DINOv2 creates transferable features from 142M self-supervised images \cite{dinov2}. DINO is a vision transformer model \cite{dino}. MAE reconstructs masked patches \cite{mae}, SimCLR maximises agreement between positive augmented views of the same image \cite{simclr} and CLIP is trained with image text supervision \cite{clip}. These approaches are not compared to in this work. DINOv2 and CLIP have already been adapted to face identification using LoRA \cite{froundation2025}. Therefore applying LoRA to visual face encoders is not novel. General and face specific models were compared by Sony et al. and it was found that fusion and contextual cropping is important \cite{foundationcomparison2025}. This motivates separation of preprocessing, adaptation and fusion in this diagnostic baseline model.\par
\subsection{Parameter-efficient adaptation}
LoRA allows efficient adaptation in language models by learning low rank updates without modifying the original parameters \cite{lora}. In this work, LoRA is applied to a visual encoder. Prompt tuning in vision (VPT) allows prompt tokens to be learnt for frozen vision transformers \cite{vpt}. In frozen language models, soft prompts have been researched \cite{prompt}. However, this is not tested in this work. Light adapters can be applied to frozen pre-trained backbones in vision to adapt them \cite{adaptformer}. However, in this work, LoRA is preferred as attention projections can be directly adapted, and fits well with the DINOv2 backbone.\par
\subsection{Score fusion and quality-aware gating}
Score fusion in biometrics has been heavily studied. Ross and Jain studied score fusion in matching \cite{fusionRossJain}. Scores can be processed and fused depending on their quality to utilise useful information within samples \cite{quality}. Weights in fusion can be learnt in order to depend on cues specific to each sample. In this work, a gate is learnt such that branch outputs are weighted according to branch embeddings and auxiliary slot inputs. As constant inputs were applied in this experiment, the gate is not demonstrably quality aware. However, unlike margin adaptive losses \cite{magface,adaface}, this gate is applied to the fusion of embeddings, and not the training of the classifier.\par
\section{Research Questions and Hypotheses}
RQ1: How much improvement is achieved by identity supervised LoRA adaptation compared to the frozen full face with trained projection baseline? RQ2: When is the periocular representation useful? What is lost in periocular representations alone? RQ3: Is performance improved by learned weighting compared to equal weighting? Does this improvement occur in ROF, MFR2, and RMFRD?\par
H1: Full face LoRA model will achieve better performance than frozen projection baseline. H2: Periocular adaptation will be beneficial in masks. However, information may be lost in sunglasses/neutral conditions. H3: Learned fusion will achieve better performance in certain low FAR points. These hypotheses were tested in ablation studies. Even if the expected improvements were not observed, results were still reported. Comparing AUC values and conducting calibration analysis investigates slightly different questions to hypothesis 3, therefore increased AUC values are not considered sufficient evidence of hypothesis 3.\par
\section{Methods}
\subsection{Branch representations}
Let \ensuremath{x_f} and \ensuremath{x_p} be full face and periocular data. \ensuremath{x_p} contains both eyes, eyebrows, top of nose and local context. \ensuremath{E_f} and \ensuremath{E_p} are DINOv2 encoders, and \ensuremath{P_f} and \ensuremath{P_p} are learnable projection heads. Normalised branch outputs are:\par
\begin{equation}
z_f=\operatorname{norm}(P_f(E_f(x_f))),\quad z_p=\operatorname{norm}(P_p(E_p(x_p))).
\end{equation}
Branches are the same resolution and normalised. If landmarks cannot be detected in a cropped image, this is recorded in the manifest. However, images are not replaced with full face images.\par
The gated-fusion inference architecture is shown only in Supplemental Fig. S1; the identity classifier is used only during training.\par
\subsection{Fixed and learned fusion}
The fixed fusion operation performs equal weighting.\par
\begin{equation}
z_{\mathrm{fixed}}=\operatorname{norm}(0.5z_f+0.5z_p).
\end{equation}
Within the learned fusion model, a weighting factor is learned.\par
\begin{equation}
g=\sigma\!\left(\operatorname{MLP}([z_f;z_p;q_f;q_p])\right),
\end{equation}
\ensuremath{q_f} and \ensuremath{q_p} are auxiliary descriptor slots. Within this experiment, these are kept constant. The final embedding is:
\begin{equation}
z_{\mathrm{gate}}=\operatorname{norm}(gz_f+(1-g)z_p).
\end{equation}
Although a gate is used in this model, it is considered a learned fusion approach. Four auxiliary descriptors and branch embeddings are input to the gate. However, during all new training and evaluation, the vector was kept constant [1,0,0,1]. In previous experiments, issues were found with the auxiliary inputs and key caching. However, this does not affect the key results of this experiment. Therefore, we do not claim the scalar value of the gate is calibrated with quality. Since the auxiliary input is constant, a constant bias is added in the first layer of the gate. Therefore, if the gate varies, this must be related to variations in the branch embeddings. Distribution of weights and replacement of learned weights explores this relationship.\par
\subsection{Training objective}
The LoRA parameters are added to certain attention projection layers. The encoder parameters are kept fixed during training. The LoRA parameters and projection/ fusion heads are optimised. For embedding \ensuremath{h_i} and training identity \ensuremath{y_i}, the per-image cross-entropy is:\par
\begin{equation}
\ell_i=-\log p(y_i\mid h_i).
\end{equation}
Only training identities define the classifier classes. EERs and retrospective TARs are calculated using the ROC curves. Thresholds are chosen based on identities from the validation set. Once thresholds have been chosen they remain fixed. The observed test TAR and FAR values are given. The score for verifying two normalised embeddings (i,j) is:\par
\begin{equation}
s(i,j)=z_i^{\top}z_j.
\end{equation}
\subsection{Model configurations}
Ablations are performed to analyse adaptation, region selection and fusion in the models. In M0, the projection has been trained but encoder is fixed.\par
\begin{table}[t]\centering
\caption{Model Comparison}
\label{tab:checked1}
\footnotesize\setlength{\tabcolsep}{2pt}
\sbox{\uploadedtablebox}{\begin{tabular}{ll}\toprule
Model & Comparison\\\midrule
M0 & Frozen encoder, trained projection Baseline heads\\
M1 & Full face DINOv2 + LoRA Adaptation\\
M2 & Periocular DINOv2 + LoRA Periocular only\\
M3 & Two LoRA, fixed fusion No gate\\
M4 & Two LoRA, learned gate Fusion\\
\bottomrule\end{tabular}}
\ifdim\wd\uploadedtablebox>\columnwidth\resizebox{\columnwidth}{!}{\usebox{\uploadedtablebox}}\else\usebox{\uploadedtablebox}\fi
\end{table}
\noindent\textit{Ethics and informed consent.}
No ethics committee reviewed this project, and no formal approval or exemption is claimed. The study reused existing ROF, MFR2, and RMFRD datasets downloaded from public repositories. The reviewed records do not establish individual participant consent or complete image-use permissions.\par
\section{Datasets and Protocol}
\subsection{ROF content and identity audit}
There were 7751 indexed files, 399 condition-specific labels, 182 normalised identity names and 6491 unique hashes. A resource illustration previously indexed as an identity was removed. The number of images in this dataset does not match with published data. All 437 annotated pickles match the blob hashes in the git repository. However, no log of image download history could be found. As such it cannot be confirmed that all images downloaded are valid or that all identities have semantically meaningful labels.\par
Suffixes attached to conditions were normalised before grouping by identity. From the original complete manifest, there were 7295 rows. There were 4 duplicated rows between training and validation data, with differing identity names. On further perceptual analysis, a near duplicate image with differing names was found between training and testing data. Despite having differing file hashes, this image differed slightly in content. In the new manifest, exact duplicates, unreadable content, conflicting labels, and both copies of the confirmed near duplicate image between splits have been removed. The new manifest contains 6171 rows (4405 training, 844 validation, 922 test). Within training, validation and testing, there are 128, 26, and 28 normalised identity labels respectively. File hashes were confirmed to be unique and identity labels disjoint before training. The manifest hash was recorded in all new checkpoints. Historical checkpoints were not continued within this experiment.\par
Each eligible identity and condition had one neutral enrolment image. Self-image comparisons were excluded. 423 comparisons were made in the masked testing protocol, with 47 genuine comparisons, and 376 impostor comparisons. 9 eligible identities were present. Within the sunglasses testing protocol, 6648 comparisons were made with 277 genuine comparisons, and 6371 impostor comparisons. 24 eligible identities in the sunglasses protocol. Within the neutral testing protocol, 15960 comparisons were made, with 570 genuine comparisons, and 15390 impostor comparisons. 28 eligible identities in the neutral protocol. The same comparisons were made in all models. In validation, comparisons were made only with images belonging to validation identities. 300 masked comparisons, 5148 sunglasses comparisons, and 14404 neutral comparisons. Within masked validation, 30 genuine comparisons and 270 impostor comparisons were made with 10 eligible identities. Due to few identities within validation data, generalisations of results are difficult to make.\par
\subsection{External protocols after recorded exclusions}
The official MFR2 protocol contains 848 pairs; 820 complete-case comparisons remain after 28 crop-endpoint exclusions. One MFR2 name (Joe Biden) exactly matched one name in ROF’s training data (18 comparisons). Removing these comparisons leaves 802 comparisons (406 genuine, 396 impostors) in this evaluation. Therefore, no known exact-name overlap remains. However, it is unknown whether aliasing, unknown identities, or overlap with pre-training has occurred. No images from MFR2 were used in new gradient updates or thresholding.\par
In the source index of RMFRD, 91849 entries exist. 22554 comparisons were made in exploratory verification. 756 comparisons were affected by exact-content conflicts across nominal identities (36 groupings). Removing these means that there are 21798 comparisons left in this protocol (1038 genuine, 20760 impostors). Therefore, this is an exact content excluded exploratory verification protocol. It is not a benchmark of independently verified semantically meaningful identities. Perceptual duplicates and mislabelled images may still exist. Since a constant auxiliary vector was used for all protocols, it cannot be claimed that a lack of quality metadata in external datasets indicates poor image quality.\par
\subsection{Metrics, calibration, and uncertainty}
Performance metrics have been given as AUC, EER, and TAR (FAR \ensuremath{\leq} \ensuremath{10^{-2}} and \ensuremath{10^{-3}} ). These metrics have been calculated on test data and do not reflect deployment thresholds. All values have been recalculated from raw scores with 10 decimal places of precision. Minimum FAR (non-zero) on masked data: \ensuremath{1/376\approx0.00266}. Therefore, the \ensuremath{10^{-3}} summary only covers a portion of the finite ROC curve with no false accepts. It does not imply a population FAR below \ensuremath{10^{-3}}.\par
Thresholds were chosen to have at most \ensuremath{\lfloor\alpha N_{\mathrm{val}}^{-}\rfloor} false accepts in validation data. In the case of ties, the next highest representable threshold was used. The threshold is then fixed in the corresponding testing conditions for ROF. MFR2 and RMFRD data were tested with the masked validation threshold derived from ROF data (external calibration not performed). The true number of false/true acceptance, FAR and TAR are given. A validation target does not establish the FAR achieved on the test set. With a desired FAR of \ensuremath{10^{-3}} in masked validation, 0 false accepts in validation are allowed (\ensuremath{N_{\mathrm{val}}^{-}=270}).\par
The variance in results from random seeds can be seen by calculating the mean and sample standard deviation in results from 3 independent seeds (7, 42, 123). Sensitivity was analysed using 1000 bootstrap samples. Multinomial sampling was carried out on identity nodes (seed 42). In a bootstrap sample, if identity i is sampled ci times, edges between identities of the same identity will be weighted by ci. Edges between impostor identities i and j will be weighted by cicj. This accounts for identities being sampled multiple times on either side of an edge. In pairwise comparisons, edge weights were applied identically to both systems. Weighting was applied to all 3 fixed seed trainings. Differences in AUC were taken as averages for each bootstrap sample. This gives a percentile range depending on trained checkpoints and selected identities. This range does not indicate uncertainty in all possible training combinations or guarantees to population security. 10 comparisons are given. No claims of significant differences due to multiple comparisons have been made.\par
\section{Experimental Settings}
\subsection{Model settings}
Cropped face and periocular modalities were scaled to 224x224. Normalisation was applied using ImageNet channel statistics. LoRA was applied to the ViT-B/14 DINOv2 backbone (projection dimension = 768). LoRA parameters: $r=8$, $\alpha=16$, dropout = 0.1. LoRA updates were applied to QKV projections. Base model parameters were fixed.\par
\subsection{Training settings}
All 15 models were trained on the CPU. 8 threads of PyTorch. 5 full epochs. Batch size = 16, AdamW optimiser, learning rate = 1e-4, weight decay = 1e-4. Cross Entropy loss. 128 labels from training were used within loss calculation. During each epoch, 4405 samples were sampled with replacement (276 batches). 1380 optimisation steps were performed. With research runs, no maximum batch counts were imposed. Classifier heads trained during training were not considered during verification. Final checkpoints from epoch 5 were used. Best epochs were not selected based on test performance.\par
\subsection{Execution environment}
Execution environment: PyTorch (version 2.14.0+cpu), deterministic. DINOv2 was cached.\par
\section{Results}
All results given below are from 15 newly trained models under the deduplication manifest. Historical numerical files are kept for auditing, and were not combined with new runs. Differences in new results from previous full case results and post-hoc filtered results are highlighted.\par
\subsection{Masked ROF verification}
The best results for masked ROF are achieved with periocular LoRA (mean AUC 0.9221). Fixed fusion models give (mean AUC 0.9134) and gated models (mean AUC 0.8989). Full face LoRA performs better than frozen projection baseline models (mean AUC 0.8852 vs 0.8303). This suggests that adaptation improves performance, and that focussing on upper facial features may be beneficial when lower regions are occluded. However, gated models do not perform best in the case of periocular only. The mean difference in performance between gated models and fixed fusion models is -0.0145. The 95\% sensitivity interval from identity resampling is -0.0560 to 0.0144. Only 9 identities exist in the masked testing, meaning a large sensitivity interval. Conditional comparisons can be made.\par
Values in low FAR columns should be treated carefully. For example, even if 0 false acceptances are observed at a test ROC point, TAR may not be 0. It may not be possible to achieve operating points below the first non-zero FAR. In seed=42, operating points with 0 FAR have been omitted. However, these have been included in metric calculation.\par
\begin{table}[t]\centering
\caption{New results from masked ROF. 47 genuine comparisons, 376 impostor comparisons. Mean values given with standard deviation in subscript. 3 seeds. TAR values represent a summary from ROC curves. They do not represent calibrated test rates from validation.}
\label{tab:checked3}
\footnotesize\setlength{\tabcolsep}{2pt}
\sbox{\uploadedtablebox}{\begin{tabular}{lllll}\toprule
Configuration & AUC & EER & TAR 0.01 & TAR 0.001\\\midrule
Frozen + Projection & \shortstack{0.8303\\\ensuremath{\pm}0.024} & \shortstack{0.2580\\\ensuremath{\pm}0.035} & \shortstack{0.1348\\\ensuremath{\pm}0.033} & \shortstack{0.0000\\\ensuremath{\pm}0.000}\\
Whole LoRA & \shortstack{0.8852\\\ensuremath{\pm}0.005} & \shortstack{0.1711\\\ensuremath{\pm}0.004} & \shortstack{0.2553\\\ensuremath{\pm}0.043} & \shortstack{0.0780\\\ensuremath{\pm}0.012}\\
Periocular LoRA & \shortstack{0.9221\\\ensuremath{\pm}0.005} & \shortstack{0.1339\\\ensuremath{\pm}0.009} & \shortstack{0.0851\\\ensuremath{\pm}0.056} & \shortstack{0.0142\\\ensuremath{\pm}0.012}\\
Fixed Fusion & \shortstack{0.9134\\\ensuremath{\pm}0.007} & \shortstack{0.1507\\\ensuremath{\pm}0.019} & \shortstack{0.2908\\\ensuremath{\pm}0.033} & \shortstack{0.1418\\\ensuremath{\pm}0.044}\\
Gating & \shortstack{0.8989\\\ensuremath{\pm}0.002} & \shortstack{0.1711\\\ensuremath{\pm}0.020} & \shortstack{0.3191\\\ensuremath{\pm}0.056} & \shortstack{0.1560\\\ensuremath{\pm}0.054}\\
\bottomrule\end{tabular}}
\ifdim\wd\uploadedtablebox>\columnwidth\resizebox{\columnwidth}{!}{\usebox{\uploadedtablebox}}\else\usebox{\uploadedtablebox}\fi
\end{table}
\begin{figure}[t]\centering
\begin{tikzpicture}
\begin{axis}[
width=\columnwidth,height=0.76\columnwidth,
font=\scriptsize,label style={font=\scriptsize},
tick label style={font=\scriptsize},title style={font=\scriptsize},
legend style={font=\tiny,draw=black!30,fill=white,inner sep=2pt},
grid=major,grid style={black!12},
title={Duplicate-excluded ROF masked; seed 42},xmode=log,xmin=0.0026595744680851063,xmax=1,ymin=0,ymax=1.01,
xlabel={Empirical false acceptance rate},ylabel={True acceptance rate},
legend pos=south east
]
\addplot[draw=plotcolor0,line width=0.65pt,no marks] coordinates {
(0.0026595744680851063,0)
(0.0053191489361702126,0)
(0.0053191489361702126,0.10638297872340426)
(0.010638297872340425,0.10638297872340426)
(0.010638297872340425,0.19148936170212766)
(0.015957446808510637,0.19148936170212766)
(0.015957446808510637,0.21276595744680851)
(0.021276595744680851,0.21276595744680851)
(0.021276595744680851,0.23404255319148937)
(0.026595744680851064,0.23404255319148937)
(0.026595744680851064,0.25531914893617019)
(0.029255319148936171,0.25531914893617019)
(0.029255319148936171,0.2978723404255319)
(0.050531914893617018,0.2978723404255319)
(0.050531914893617018,0.31914893617021278)
(0.066489361702127658,0.31914893617021278)
(0.066489361702127658,0.34042553191489361)
(0.069148936170212769,0.34042553191489361)
(0.069148936170212769,0.36170212765957449)
(0.082446808510638292,0.36170212765957449)
(0.082446808510638292,0.38297872340425532)
(0.10106382978723404,0.38297872340425532)
(0.10106382978723404,0.40425531914893614)
(0.11170212765957446,0.40425531914893614)
(0.11170212765957446,0.42553191489361702)
(0.11968085106382979,0.42553191489361702)
(0.11968085106382979,0.44680851063829785)
(0.12234042553191489,0.44680851063829785)
(0.12234042553191489,0.46808510638297873)
(0.14361702127659576,0.46808510638297873)
(0.14361702127659576,0.48936170212765956)
(0.15425531914893617,0.48936170212765956)
(0.15425531914893617,0.51063829787234039)
(0.15691489361702127,0.51063829787234039)
(0.15691489361702127,0.53191489361702127)
(0.16489361702127658,0.53191489361702127)
(0.16489361702127658,0.55319148936170215)
(0.18351063829787234,0.55319148936170215)
(0.18351063829787234,0.57446808510638303)
(0.18617021276595744,0.57446808510638303)
(0.18617021276595744,0.5957446808510638)
(0.2047872340425532,0.5957446808510638)
(0.2047872340425532,0.61702127659574468)
(0.22074468085106383,0.61702127659574468)
(0.22074468085106383,0.63829787234042556)
(0.22872340425531915,0.63829787234042556)
(0.22872340425531915,0.68085106382978722)
(0.23670212765957446,0.68085106382978722)
(0.23670212765957446,0.74468085106382975)
(0.24202127659574468,0.74468085106382975)
(0.24202127659574468,0.76595744680851063)
(0.24468085106382978,0.76595744680851063)
(0.24468085106382978,0.78723404255319152)
(0.26861702127659576,0.78723404255319152)
(0.26861702127659576,0.80851063829787229)
(0.28723404255319152,0.80851063829787229)
(0.28723404255319152,0.82978723404255317)
(0.30585106382978722,0.82978723404255317)
(0.30585106382978722,0.85106382978723405)
(0.31117021276595747,0.85106382978723405)
(0.31117021276595747,0.87234042553191493)
(0.32978723404255317,0.87234042553191493)
(0.32978723404255317,0.91489361702127658)
(0.35904255319148937,0.91489361702127658)
(0.35904255319148937,0.93617021276595747)
(0.39627659574468083,0.93617021276595747)
(0.39627659574468083,0.95744680851063835)
(0.43617021276595747,0.95744680851063835)
(0.43617021276595747,0.97872340425531912)
(0.47606382978723405,0.97872340425531912)
(0.47606382978723405,1)
(1,1)
};
\addlegendentry{Frozen + projection (0.841)}
\addplot[draw=plotcolor1,line width=0.65pt,no marks] coordinates {
(0.0026595744680851063,0.063829787234042548)
(0.0026595744680851063,0.14893617021276595)
(0.0053191489361702126,0.14893617021276595)
(0.0053191489361702126,0.1702127659574468)
(0.0079787234042553185,0.1702127659574468)
(0.0079787234042553185,0.21276595744680851)
(0.010638297872340425,0.21276595744680851)
(0.010638297872340425,0.40425531914893614)
(0.029255319148936171,0.40425531914893614)
(0.029255319148936171,0.42553191489361702)
(0.058510638297872342,0.42553191489361702)
(0.058510638297872342,0.46808510638297873)
(0.061170212765957445,0.46808510638297873)
(0.061170212765957445,0.48936170212765956)
(0.063829787234042548,0.48936170212765956)
(0.063829787234042548,0.55319148936170215)
(0.074468085106382975,0.55319148936170215)
(0.074468085106382975,0.57446808510638303)
(0.077127659574468085,0.57446808510638303)
(0.077127659574468085,0.5957446808510638)
(0.087765957446808512,0.5957446808510638)
(0.087765957446808512,0.61702127659574468)
(0.090425531914893623,0.61702127659574468)
(0.090425531914893623,0.63829787234042556)
(0.093085106382978719,0.63829787234042556)
(0.093085106382978719,0.7021276595744681)
(0.10106382978723404,0.7021276595744681)
(0.10106382978723404,0.72340425531914898)
(0.13031914893617022,0.72340425531914898)
(0.13031914893617022,0.76595744680851063)
(0.13829787234042554,0.76595744680851063)
(0.13829787234042554,0.78723404255319152)
(0.14627659574468085,0.78723404255319152)
(0.14627659574468085,0.82978723404255317)
(0.19680851063829788,0.82978723404255317)
(0.19680851063829788,0.85106382978723405)
(0.21276595744680851,0.85106382978723405)
(0.21276595744680851,0.87234042553191493)
(0.23404255319148937,0.87234042553191493)
(0.23404255319148937,0.8936170212765957)
(0.23670212765957446,0.8936170212765957)
(0.23670212765957446,0.91489361702127658)
(0.53457446808510634,0.91489361702127658)
(0.53457446808510634,0.93617021276595747)
(0.58776595744680848,0.93617021276595747)
(0.58776595744680848,0.95744680851063835)
(0.8271276595744681,0.95744680851063835)
(0.8271276595744681,0.97872340425531912)
(0.88297872340425532,0.97872340425531912)
(0.88297872340425532,1)
(1,1)
};
\addlegendentry{Whole LoRA (0.880)}
\addplot[draw=plotcolor2,line width=0.65pt,no marks] coordinates {
(0.0026595744680851063,0.021276595744680851)
(0.0026595744680851063,0.042553191489361701)
(0.010638297872340425,0.042553191489361701)
(0.010638297872340425,0.063829787234042548)
(0.015957446808510637,0.063829787234042548)
(0.015957446808510637,0.10638297872340426)
(0.018617021276595744,0.10638297872340426)
(0.018617021276595744,0.1276595744680851)
(0.021276595744680851,0.1276595744680851)
(0.021276595744680851,0.14893617021276595)
(0.023936170212765957,0.14893617021276595)
(0.023936170212765957,0.23404255319148937)
(0.026595744680851064,0.23404255319148937)
(0.026595744680851064,0.25531914893617019)
(0.034574468085106384,0.25531914893617019)
(0.034574468085106384,0.38297872340425532)
(0.037234042553191488,0.38297872340425532)
(0.037234042553191488,0.44680851063829785)
(0.039893617021276598,0.44680851063829785)
(0.039893617021276598,0.48936170212765956)
(0.047872340425531915,0.48936170212765956)
(0.047872340425531915,0.51063829787234039)
(0.053191489361702128,0.51063829787234039)
(0.053191489361702128,0.57446808510638303)
(0.055851063829787231,0.57446808510638303)
(0.055851063829787231,0.68085106382978722)
(0.058510638297872342,0.68085106382978722)
(0.058510638297872342,0.7021276595744681)
(0.063829787234042548,0.7021276595744681)
(0.063829787234042548,0.72340425531914898)
(0.066489361702127658,0.72340425531914898)
(0.066489361702127658,0.76595744680851063)
(0.071808510638297879,0.76595744680851063)
(0.071808510638297879,0.78723404255319152)
(0.079787234042553196,0.78723404255319152)
(0.079787234042553196,0.80851063829787229)
(0.10904255319148937,0.80851063829787229)
(0.10904255319148937,0.82978723404255317)
(0.11436170212765957,0.82978723404255317)
(0.11436170212765957,0.85106382978723405)
(0.13031914893617022,0.85106382978723405)
(0.13031914893617022,0.87234042553191493)
(0.14627659574468085,0.87234042553191493)
(0.14627659574468085,0.8936170212765957)
(0.17553191489361702,0.8936170212765957)
(0.17553191489361702,0.91489361702127658)
(0.21542553191489361,0.91489361702127658)
(0.21542553191489361,0.93617021276595747)
(0.22340425531914893,0.93617021276595747)
(0.22340425531914893,0.95744680851063835)
(0.38563829787234044,0.95744680851063835)
(0.38563829787234044,0.97872340425531912)
(0.48138297872340424,0.97872340425531912)
(0.48138297872340424,1)
(1,1)
};
\addlegendentry{Periocular LoRA (0.926)}
\addplot[draw=plotcolor3,line width=0.65pt,no marks] coordinates {
(0.0026595744680851063,0.19148936170212766)
(0.0026595744680851063,0.27659574468085107)
(0.0053191489361702126,0.27659574468085107)
(0.0053191489361702126,0.2978723404255319)
(0.010638297872340425,0.2978723404255319)
(0.010638297872340425,0.31914893617021278)
(0.023936170212765957,0.31914893617021278)
(0.023936170212765957,0.36170212765957449)
(0.031914893617021274,0.36170212765957449)
(0.031914893617021274,0.40425531914893614)
(0.045212765957446811,0.40425531914893614)
(0.045212765957446811,0.42553191489361702)
(0.047872340425531915,0.42553191489361702)
(0.047872340425531915,0.46808510638297873)
(0.050531914893617018,0.46808510638297873)
(0.050531914893617018,0.53191489361702127)
(0.058510638297872342,0.53191489361702127)
(0.058510638297872342,0.5957446808510638)
(0.061170212765957445,0.5957446808510638)
(0.061170212765957445,0.65957446808510634)
(0.063829787234042548,0.65957446808510634)
(0.063829787234042548,0.7021276595744681)
(0.069148936170212769,0.7021276595744681)
(0.069148936170212769,0.74468085106382975)
(0.090425531914893623,0.74468085106382975)
(0.090425531914893623,0.76595744680851063)
(0.10638297872340426,0.76595744680851063)
(0.10638297872340426,0.78723404255319152)
(0.12234042553191489,0.78723404255319152)
(0.12234042553191489,0.80851063829787229)
(0.125,0.80851063829787229)
(0.125,0.82978723404255317)
(0.1276595744680851,0.82978723404255317)
(0.1276595744680851,0.85106382978723405)
(0.13297872340425532,0.85106382978723405)
(0.13297872340425532,0.8936170212765957)
(0.32446808510638298,0.8936170212765957)
(0.32446808510638298,0.91489361702127658)
(0.33510638297872342,0.91489361702127658)
(0.33510638297872342,0.93617021276595747)
(0.46010638297872342,0.93617021276595747)
(0.46010638297872342,0.95744680851063835)
(0.5292553191489362,0.95744680851063835)
(0.5292553191489362,0.97872340425531912)
(0.5957446808510638,0.97872340425531912)
(0.5957446808510638,1)
(1,1)
};
\addlegendentry{Fixed fusion (0.912)}
\addplot[draw=plotcolor4,line width=0.65pt,no marks] coordinates {
(0.0026595744680851063,0.14893617021276595)
(0.0026595744680851063,0.25531914893617019)
(0.0053191489361702126,0.25531914893617019)
(0.0053191489361702126,0.2978723404255319)
(0.0079787234042553185,0.2978723404255319)
(0.0079787234042553185,0.38297872340425532)
(0.018617021276595744,0.38297872340425532)
(0.018617021276595744,0.40425531914893614)
(0.031914893617021274,0.40425531914893614)
(0.031914893617021274,0.44680851063829785)
(0.034574468085106384,0.44680851063829785)
(0.034574468085106384,0.53191489361702127)
(0.042553191489361701,0.53191489361702127)
(0.042553191489361701,0.57446808510638303)
(0.050531914893617018,0.57446808510638303)
(0.050531914893617018,0.5957446808510638)
(0.053191489361702128,0.5957446808510638)
(0.053191489361702128,0.63829787234042556)
(0.061170212765957445,0.63829787234042556)
(0.061170212765957445,0.65957446808510634)
(0.063829787234042548,0.65957446808510634)
(0.063829787234042548,0.68085106382978722)
(0.077127659574468085,0.68085106382978722)
(0.077127659574468085,0.7021276595744681)
(0.079787234042553196,0.7021276595744681)
(0.079787234042553196,0.72340425531914898)
(0.11170212765957446,0.72340425531914898)
(0.11170212765957446,0.74468085106382975)
(0.11702127659574468,0.74468085106382975)
(0.11702127659574468,0.76595744680851063)
(0.1276595744680851,0.76595744680851063)
(0.1276595744680851,0.78723404255319152)
(0.17819148936170212,0.78723404255319152)
(0.17819148936170212,0.80851063829787229)
(0.18882978723404256,0.80851063829787229)
(0.18882978723404256,0.82978723404255317)
(0.21542553191489361,0.82978723404255317)
(0.21542553191489361,0.85106382978723405)
(0.23404255319148937,0.85106382978723405)
(0.23404255319148937,0.87234042553191493)
(0.25797872340425532,0.87234042553191493)
(0.25797872340425532,0.8936170212765957)
(0.28989361702127658,0.8936170212765957)
(0.28989361702127658,0.91489361702127658)
(0.37234042553191488,0.91489361702127658)
(0.37234042553191488,0.93617021276595747)
(0.53191489361702127,0.93617021276595747)
(0.53191489361702127,0.95744680851063835)
(0.64627659574468088,0.95744680851063835)
(0.64627659574468088,0.97872340425531912)
(0.74202127659574468,0.97872340425531912)
(0.74202127659574468,1)
(1,1)
};
\addlegendentry{Learned gating (0.898)}
\end{axis}
\end{tikzpicture}
\caption{New masked ROF ROC (seed=42). Only positive empirical FAR values are shown. The first non-zero FAR value is 1/376. Operating points with zero FAR have been excluded from this log scale graph.}
\label{fig:checked1}
\end{figure}
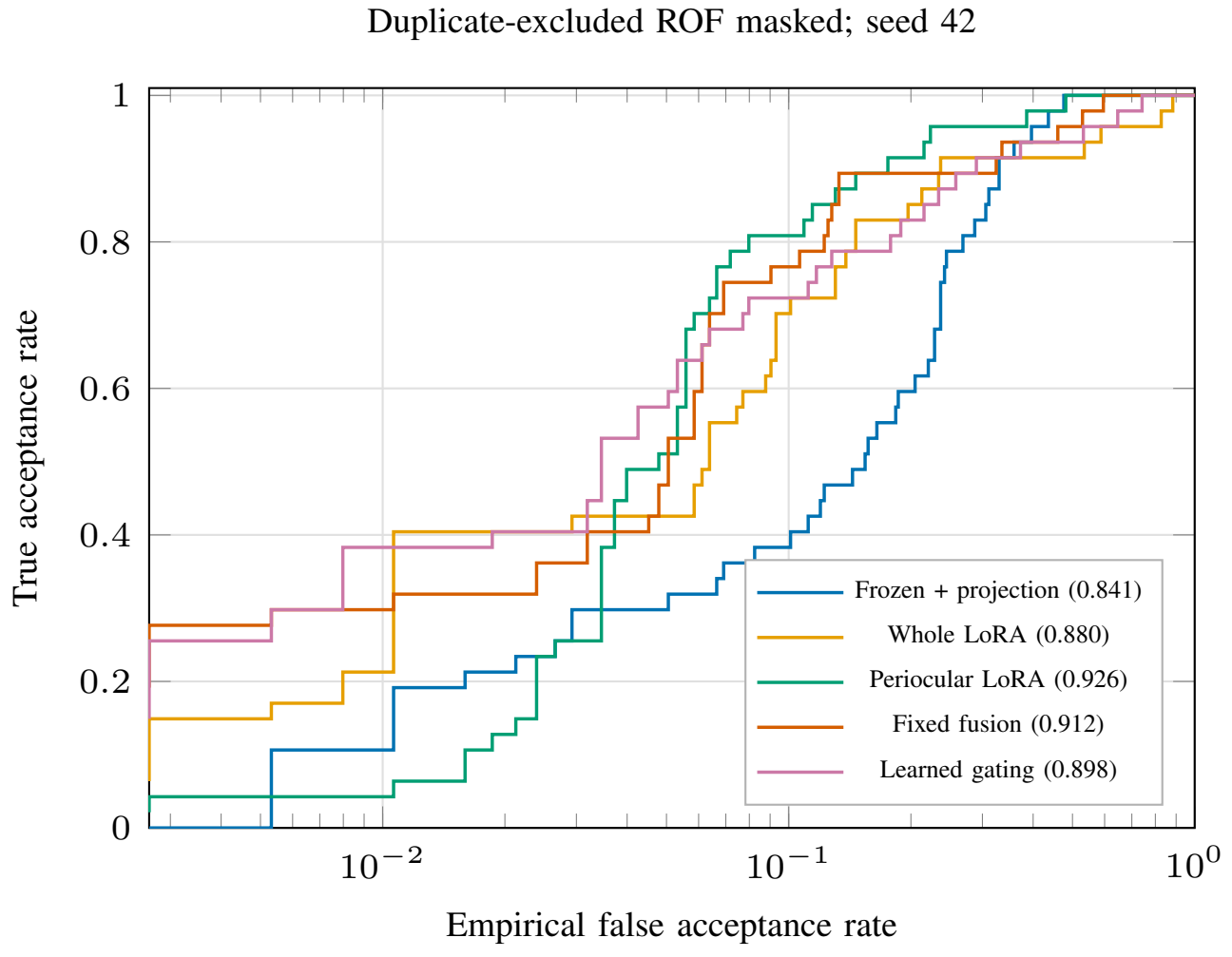
As a post-hoc exploratory cohort-composition sensitivity analysis, each of the nine masked test identities was removed in turn, together with its incident comparisons, without retraining. For 8/9 subsets, periocular LoRA gave the best mean AUC.\par
Fixed fusion gives the best results in the remaining subset. The average difference between the gated and fixed fusion is negative across all subsets (-0.0243 to -0.0026). This describes sensitivity within the observed cohort; it does not establish population bias or generalisation to unseen cohorts.\par
\subsection{Sunglasses and neutral ROF verification}
With sunglasses the periocular only model performs poorly, with a mean AUC of 0.7905 compared to frozen full face (0.8780). Fixed fusion performs best (mean AUC 0.8950), closely followed by gated (0.8924) and full face LoRA (0.8913). The average difference between the gated and fixed models is -0.0026. The 95\% sensitivity interval is therefore [-0.0099, 0.0061]. Thus, there is not guaranteed improvement in performance by adding extra branches and learning a weighting combination.\par
With neutral images, learned gate fusion gives the best average AUC (0.9458) compared to fixed fusion (0.9428) and full face LoRA (0.9413). The average difference between gated and fixed fusion is 0.003. The 95\% sensitivity interval is [-0.0046, 0.0106]. Due to the small difference, there is not conclusive evidence that population performance is better in gated models. The periocular only model had a mean AUC of 0.8875. Thus, some information was lost by removing clean lower face cues. Including a neutral control in Table~\ref{tab:checked5} ensures that claims for different conditions cannot be solely made from probed images with occlusions.\par
\begin{table}[t]\centering
\caption{Results for new sunglasses ROF data. (277 Genuine and 6371 Impostors). AUC/EER/TAR values are averaged across sample seeds (standard deviation in subscript), averages taken over 3 seeds. TAR values have been derived from the ROC curves, and are not calibrated test rates.}
\label{tab:checked4}
\footnotesize\setlength{\tabcolsep}{2pt}
\sbox{\uploadedtablebox}{
\begin{tabular}{lllll}\toprule
Config & AUC & EER & TAR 0.01 & TAR 0.001\\\midrule
Frozen + Projection & \shortstack{0.8780\\\ensuremath{\pm}0.003} & \shortstack{0.1774\\\ensuremath{\pm}0.004} & \shortstack{0.1095\\\ensuremath{\pm}0.002} & \shortstack{0.0132\\\ensuremath{\pm}0.004}\\
Whole LoRA & \shortstack{0.8913\\\ensuremath{\pm}0.002} & \shortstack{0.1596\\\ensuremath{\pm}0.005} & \shortstack{0.1661\\\ensuremath{\pm}0.010} & \shortstack{0.0337\\\ensuremath{\pm}0.004}\\
Periocular LoRA & \shortstack{0.7905\\\ensuremath{\pm}0.014} & \shortstack{0.2632\\\ensuremath{\pm}0.007} & \shortstack{0.0710\\\ensuremath{\pm}0.010} & \shortstack{0.0048\\\ensuremath{\pm}0.004}\\
Fixed Fusion & \shortstack{0.8950\\\ensuremath{\pm}0.002} & \shortstack{0.1531\\\ensuremath{\pm}0.008} & \shortstack{0.1564\\\ensuremath{\pm}0.012} & \shortstack{0.0144\\\ensuremath{\pm}0.006}\\
Gate Fusion & \shortstack{0.8924\\\ensuremath{\pm}0.006} & \shortstack{0.1544\\\ensuremath{\pm}0.001} & \shortstack{0.1384\\\ensuremath{\pm}0.016} & \shortstack{0.0373\\\ensuremath{\pm}0.008}\\
\bottomrule\end{tabular}}
\ifdim\wd\uploadedtablebox>\columnwidth\resizebox{\columnwidth}{!}{\usebox{\uploadedtablebox}}\else\usebox{\uploadedtablebox}\fi
\end{table}
\begin{table}[t]\centering
\caption{Results for new neutral ROF data. (570 Genuine and 15390 Impostors). AUC/EER/TAR values are averaged across sample seeds (standard deviation in subscript), averages taken over 3 seeds. TAR values have been derived from the ROC curves, and are not calibrated test rates.}
\label{tab:checked5}
\footnotesize\setlength{\tabcolsep}{2pt}
\sbox{\uploadedtablebox}{%
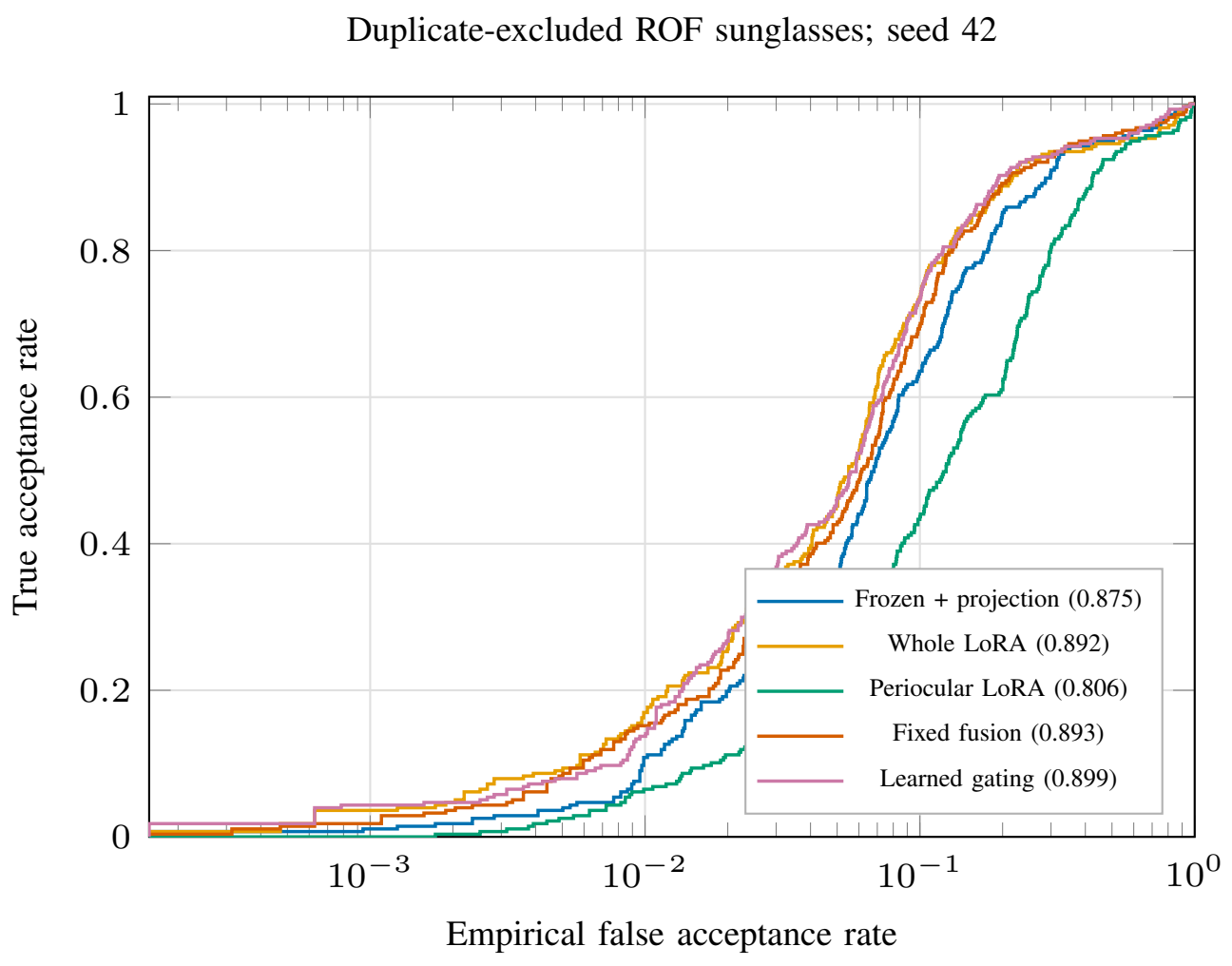
\caption{New set of sunglasses (Seed 42). Only positive empirical FAR values are displayed. All systems use the same frozen pairs.}
\label{fig:checked2}
\end{figure}
\subsection{Known-overlap-excluded MFR2}
The best average AUC value for learned fusion is achieved at 0.9548, evaluated on 802 pairs and averaged over three seeds. The average AUC for fixed fusion was 0.9398, meaning an average improvement in AUC of 0.015 was achieved by using learned fusion over fixed fusion. The 95\% identity-resampling sensitivity interval for this improvement is [0.0044, 0.0283]. For full face LoRA models, an average AUC was achieved of 0.9360, which shows improvement over the frozen baseline model (0.8412). This improvement is larger than that in gate fusions over fixed fusions. No labels in the target domain were used in gradient updates or calibration.\par
After excluding the known exact-name overlap, the retained 802-pair results differ from the original 820-pair analysis. No known normalised exact-name overlap with ROF training remains; unknown aliases and overlap with DINOv2 pretraining cannot be ruled out. Ordered performance under known protocols can be seen from ROC results in external data. Individual thresholds from ROF validation data have been applied in order to calculate performance.\par
\begin{table}[t]\centering
\caption{Excluded known name overlap MFR2. (406 Genuine and 396 Impostors). AUC/EER/TAR values are averaged across sample seeds (standard deviation in subscript), averages taken over 3 seeds. TAR values have been derived from the ROC curves, and are not calibrated test rates.}
\label{tab:checked6}
\footnotesize\setlength{\tabcolsep}{2pt}
\sbox{\uploadedtablebox}{

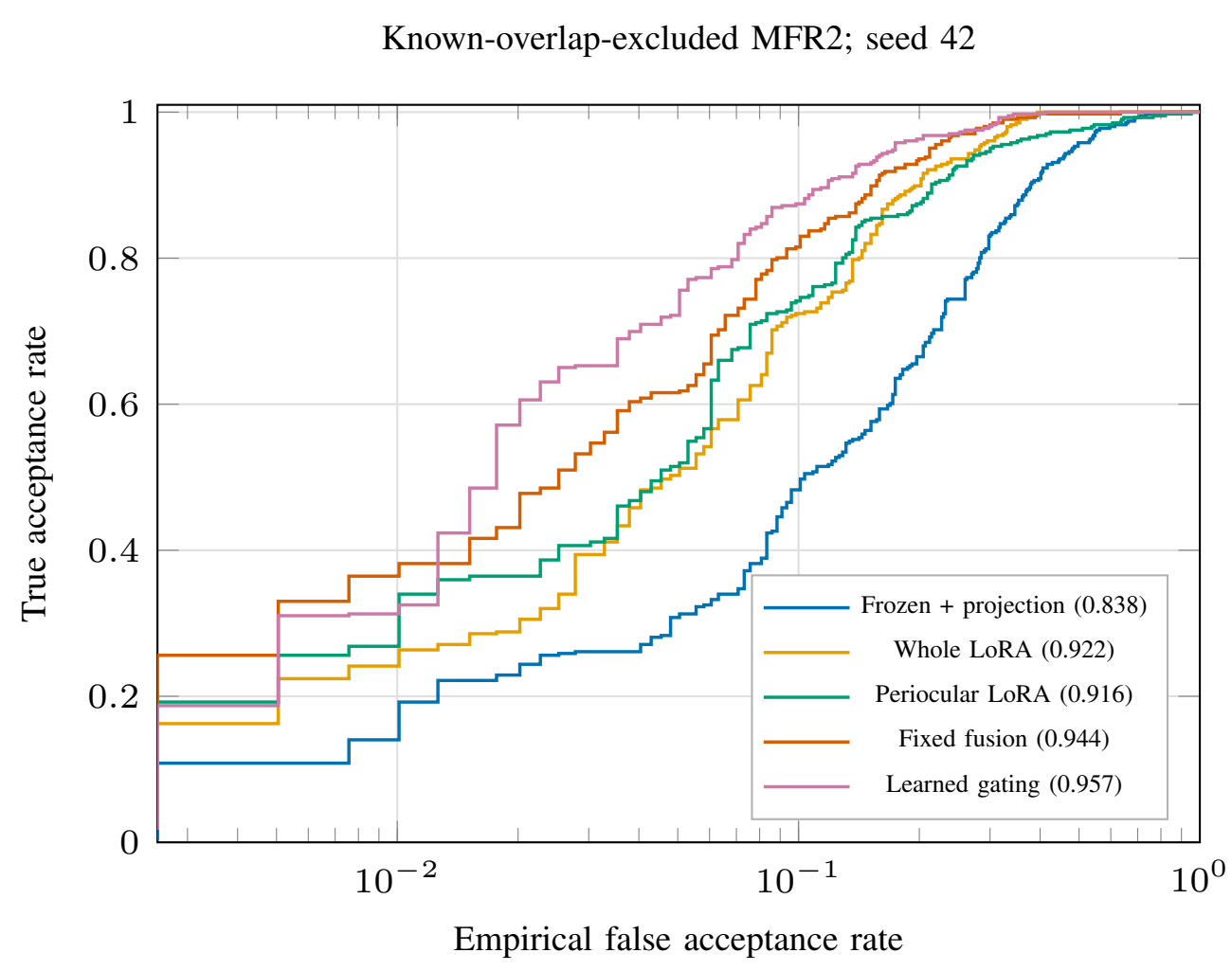
\caption{MFR2 ROC excluding known name overlap (seed 42). Only positive empirical FAR values included. Auxiliary input is the same as in training.}
\label{fig:checked3}
\end{figure}
\subsection{Exact-conflict-excluded RMFRD}
The best exploratory results come from fixed fusion with a mean AUC of 0.7240. Next best are learned fusion (0.7175) and then periocular LoRA (0.7132). Whole face LoRA performs with a mean AUC of 0.7045 and frozen baseline with 0.6957. The average difference is -0.0066 between gated and fixed models. The 95\% sensitivity interval is [-0.0135, 0.0004]. As 0 is within the final descriptive interval, conclusive ordering could not be determined. RMFRD labels have not been independently validated.\par
A much lower AUC than MFR2 demonstrates difficulties in transfer. Although known exact conflicts have been removed, uncertainty within annotations and acquisition means strong claims cannot be made for unseen identities or deployments. Summary results for all 5 test scenarios are presented in Fig 4. Variability between seeds is presented, rather than optimum datasets.\par
\begin{table}[t]\centering
\caption{Exploratory results for excluded exact conflict RMFRD. (1038 Genuine and 20760 Impostors). AUC/EER/TAR values are averaged across sample seeds (standard deviation in subscript), averages taken over 3 seeds. TAR values have been derived from the ROC curves, and are not calibrated test rates.}
\label{tab:checked7}
\footnotesize\setlength{\tabcolsep}{2pt}
\sbox{\uploadedtablebox}{\begin{tabular}{lllll}\toprule
Config & AUC & EER & TAR 0.01 & TAR 0.001\\\midrule
Frozen + Projection & \shortstack{0.6957\\\ensuremath{\pm}0.004} & \shortstack{0.3518\\\ensuremath{\pm}0.002} & \shortstack{0.0267\\\ensuremath{\pm}0.004} & \shortstack{0.0035\\\ensuremath{\pm}0.001}\\
Whole LoRA & \shortstack{0.7045\\\ensuremath{\pm}0.010} & \shortstack{0.3507\\\ensuremath{\pm}0.011} & \shortstack{0.0324\\\ensuremath{\pm}0.009} & \shortstack{0.0045\\\ensuremath{\pm}0.001}\\
Periocular LoRA & \shortstack{0.7132\\\ensuremath{\pm}0.004} & \shortstack{0.3412\\\ensuremath{\pm}0.005} & \shortstack{0.0373\\\ensuremath{\pm}0.001} & \shortstack{0.0035\\\ensuremath{\pm}0.002}\\
Fixed Fusion & \shortstack{0.7240\\\ensuremath{\pm}0.010} & \shortstack{0.3358\\\ensuremath{\pm}0.008} & \shortstack{0.0459\\\ensuremath{\pm}0.003} & \shortstack{0.0071\\\ensuremath{\pm}0.002}\\
Gate Fusion & \shortstack{0.7175\\\ensuremath{\pm}0.012} & \shortstack{0.3356\\\ensuremath{\pm}0.010} & \shortstack{0.0421\\\ensuremath{\pm}0.004} & \shortstack{0.0083\\\ensuremath{\pm}0.002}\\
\bottomrule\end{tabular}}
\ifdim\wd\uploadedtablebox>\columnwidth\resizebox{\columnwidth}{!}{\usebox{\uploadedtablebox}}\else\usebox{\uploadedtablebox}\fi
\end{table}
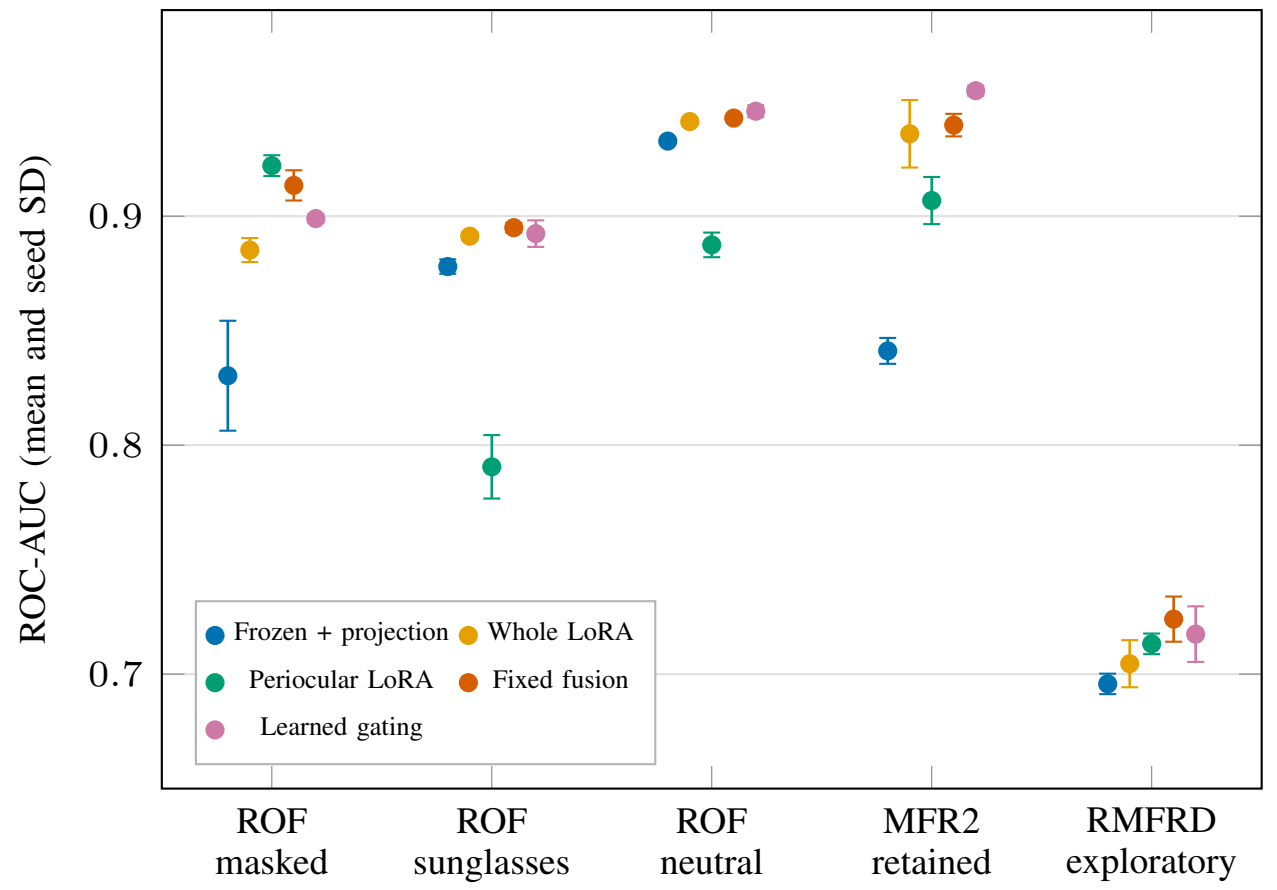
\begin{figure}[t]\centering
\begin{tikzpicture}
\begin{axis}[
width=\columnwidth,height=0.76\columnwidth,
font=\scriptsize,label style={font=\scriptsize},
tick label style={font=\scriptsize},title style={font=\scriptsize},
legend style={font=\tiny,draw=black!30,fill=white,inner sep=2pt},
grid=major,grid style={black!12},
ylabel={ROC-AUC (mean and seed SD)},ymin=0.65,ymax=0.99,xmin=-0.5,xmax=4.5,
xtick={0,1,2,3,4},
xticklabels={{\shortstack{ROF\\masked}},{\shortstack{ROF\\sunglasses}},{\shortstack{ROF\\neutral}},{\shortstack{MFR2\\retained}},{\shortstack{RMFRD\\exploratory}}},
xmajorgrids=false,legend pos=south west,legend columns=2
]
\addplot[only marks,mark=*,mark size=1.6pt,color=plotcolor0,
error bars/.cd,y dir=both,y explicit,error bar style={line width=0.5pt}]
table[x index=0,y index=1,y error index=2,row sep=\\]{
-0.20000000000000001 0.83031537649011611 0.024009389049410187\\
0.80000000000000004 0.87798918119691338 0.0031403780650324344\\
1.8 0.93278984226865624 0.0011990333379985386\\
2.7999999999999998 0.84116410409513864 0.0056618904915603655\\
3.7999999999999998 0.69572352097494938 0.004475232634236723\\
};
\addlegendentry{Frozen + projection}
\addplot[only marks,mark=*,mark size=1.6pt,color=plotcolor1,
error bars/.cd,y dir=both,y explicit,error bar style={line width=0.5pt}]
table[x index=0,y index=1,y error index=2,row sep=\\]{
-0.10000000000000001 0.8851856043458578 0.0052344983004311224\\
0.90000000000000002 0.89125722168044463 0.0017190874349601571\\
1.8999999999999999 0.94130357298921974 0.0015120691875136409\\
2.8999999999999999 0.93597303080061689 0.014745615162669317\\
3.8999999999999999 0.70453841065831113 0.010296224419906427\\
};
\addlegendentry{Whole LoRA}
\addplot[only marks,mark=*,mark size=1.6pt,color=plotcolor2,
error bars/.cd,y dir=both,y explicit,error bar style={line width=0.5pt}]
table[x index=0,y index=1,y error index=2,row sep=\\]{
0 0.92206126452391735 0.0045524452880456702\\
1 0.79053098794345089 0.013856007731609677\\
2 0.88745475340940605 0.0053837190991230357\\
3 0.90682481630757505 0.010287570017318776\\
4 0.7132280192752477 0.0044944132977217735\\
};
\addlegendentry{Periocular LoRA}
\addplot[only marks,mark=*,mark size=1.6pt,color=plotcolor3,
error bars/.cd,y dir=both,y explicit,error bar style={line width=0.5pt}]
table[x index=0,y index=1,y error index=2,row sep=\\]{
0.10000000000000001 0.91344122529047844 0.0066064406153263774\\
1.1000000000000001 0.89497215968642507 0.0021271018473859125\\
2.1000000000000001 0.94283012056891202 0.0013709951074429309\\
3.1000000000000001 0.93976712942230167 0.0049228599532323896\\
4.0999999999999996 0.72401490626581688 0.0098696936745453299\\
};
\addlegendentry{Fixed fusion}
\addplot[only marks,mark=*,mark size=1.6pt,color=plotcolor4,
error bars/.cd,y dir=both,y explicit,error bar style={line width=0.5pt}]
table[x index=0,y index=1,y error index=2,row sep=\\]{
0.20000000000000001 0.89893617021276595 0.0018835804636818851\\
1.2 0.89237427188216156 0.0057997664327799035\\
2.2000000000000002 0.94582572415444066 0.0026089246796105105\\
3.2000000000000002 0.95479424789769618 0.0022026204641751933\\
4.2000000000000002 0.71745150250654943 0.012143347834740073\\
};
\addlegendentry{Learned gating}
\end{axis}
\end{tikzpicture}
\caption{Average AUC (with single sample SD from 3 seeds) for models trained on CPU. Known name overlaps are excluded in MFR2. Known exact content conflicts are excluded in RMFRD.}
\label{fig:checked4}
\end{figure}
\subsection{Validation-calibrated operating points}
Results for models with a target False Acceptance Rate (FAR) of 0.01 in validation are presented in Table~\ref{tab:checked8}. Results show mean True Acceptance Rate (TAR) and False Acceptance Rate (FAR) in testing. Count of accepts in testing for each seed given. With ROF, the same validation scenario was used. With external data sets, masked ROF validation was used. It is possible that the test FAR may be higher than the target validation FAR. In particular, testing on neutral images results in significantly higher false acceptance rates than expected from validation. If preselected thresholds are presented in a ROC curve for testing this can be misleading.\par
For a target FAR of 0.001 in validation, zero false acceptances may be observed in validation. This does not guarantee a population FAR of 0.001. For all models, complete threshold values, and counts of accepts across calibration/testing for each seed are provided in the machine readable reports. Values given for both target FARs. Calibration is done according to a set rule, and is unchanged regardless of error in testing.\par
\begin{table}[t]\centering
\caption{Gate model with target FAR of 0.01 in validation. Averages from 3 seeds. Values for TA/FA indicate number of accepts in testing for each seed in turn. For external datasets, thresholds are transferred from masked ROF validation.}
\label{tab:checked8}
\footnotesize\setlength{\tabcolsep}{2pt}
\sbox{\uploadedtablebox}{\begin{tabular}{lllll}\toprule
Protocol & TAR & FAR & TA & FA\\\midrule
Masked & 0.3546 & 0.0222 & 16/18/16 & 13/5/7\\
Sunglasses & 0.1252 & 0.0089 & 37/27/40 & 74/46/50\\
Neutral & 0.5076 & 0.0325 & 278/285/305 & 491/424/585\\
MFR2 & 0.6814 & 0.0396 & 299/264/267 & 21/10/16\\
RMFRD & 0.6426 & 0.3230 & 765/569/667 & 8417/5424/6273\\
\bottomrule\end{tabular}}
\ifdim\wd\uploadedtablebox>\columnwidth\resizebox{\columnwidth}{!}{\usebox{\uploadedtablebox}}\else\usebox{\uploadedtablebox}\fi
\end{table}
\subsection{Same-branch gate controls}
To isolate the effect of fusion rule changing, the same checkpoints were used from the gate model. The branch features were held fixed, and the inference weighting rule was varied using equal weighting, average weighting (based on validation only) and permuted weighting within each protocol. Results for single branch features were also obtained. Average weighting was performed based on the endpoints in validation only. Permutations were performed in each protocol using seed=42, so that unique endpoints were obtained within each protocol. No cross protocol mixing occurred. This is a post-hoc analysis of weighting rules, not new training runs; the permutation is a diagnostic control.\par
\begin{table}[t]\centering
\caption{Average AUC values for fusion control, with same trained gate branch. Not separately trained M3 models.}
\label{tab:checked9}
\footnotesize\setlength{\tabcolsep}{2pt}
\sbox{\uploadedtablebox}{\begin{tabular}{lllll}\toprule
Protocol & Equal & Learned & Average & Permutation\\\midrule
Masked & 0.9250 & 0.8989 & 0.8946 & 0.8992\\
Sunglasses & 0.8854 & 0.8924 & 0.8934 & 0.8942\\
Neutral & 0.9421 & 0.9458 & 0.9464 & 0.9428\\
MFR2 & 0.9512 & 0.9548 & 0.9420 & 0.9391\\
RMFRD & 0.7126 & 0.7175 & 0.7181 & 0.7162\\
\bottomrule\end{tabular}}
\ifdim\wd\uploadedtablebox>\columnwidth\resizebox{\columnwidth}{!}{\usebox{\uploadedtablebox}}\else\usebox{\uploadedtablebox}\fi
\end{table}
For masked ROF images, using equal weighting gives higher average AUC (0.9250) than learned (0.8989). For sunglasses and neutral images slightly higher scores are obtained by using average weighting. For sunglasses images, permutation results in a slightly higher AUC than learned weighting (0.8942 vs 0.8924). However for MFR2 images, greater benefit is seen by using image-dependent weighting. Using learned weighting gives a mean AUC of 0.9548, whereas average and permutation give 0.9420 and 0.9391. Thus, evidence of benefit from sample dependency in weighting depends on test scenario.\par
Mean learned face-branch gate weights across seeds range from 0.709 to 0.749 in masked ROF, and from 0.825 to 0.841 in sunglasses. Thus, more weighting occurs in the face branch for sunglasses scenarios. However, there is variability within seeds. Therefore there may be condition-dependent weighting in branches. However weighting does not indicate reliable calibrated detection, or causative detection of occlusions. Including constant auxiliary input within the bias of the first layer only has a very small effect on gate weights (max difference between recorded checkpoints of \ensuremath{1.20\times10^{-7}}). Thus, no image-specific quality information is being conveyed through auxiliary fields.\par
\begin{figure}[t]\centering
\begin{tikzpicture}
\begin{axis}[
width=\columnwidth,height=0.76\columnwidth,
font=\scriptsize,label style={font=\scriptsize},
tick label style={font=\scriptsize},title style={font=\scriptsize},
legend style={font=\tiny,draw=black!30,fill=white,inner sep=2pt},
grid=major,grid style={black!12},
title={ROF masked},
xlabel={Cosine score},ylabel={Within-class density},ymin=0,
xmin=-0.43530228139999999,xmax=0.86573591969999997,legend pos=north west
]
\addplot[ybar interval,area legend,fill=plotcolor3,fill opacity=0.5,draw=none] coordinates {
(-0.43530228139999999,0.25552424842680393)
(-0.38326075335600002,0.45994364716824659)
(-0.33121922531199999,1.2265163924486588)
(-0.27917769726800001,1.9930891377290694)
(-0.22713616922400001,2.5041376345826767)
(-0.17509464118000001,2.0952988370997923)
(-0.12305311313600004,2.2997182358412327)
(-0.071011585092000007,1.3798309415047412)
(-0.018970057048000033,1.1243066930779371)
(0.033071470995999941,0.91988729433649319)
(0.08511299903999997,0.76657274528041175)
(0.13715452708399994,0.56215334653896853)
(0.18919605512799992,1.0220969937072157)
(0.24123758317199989,0.35773394779752471)
(0.29327911121599998,0.51104849685360787)
(0.34532063925999995,0.25552424842680393)
(0.39736216730399992,0.20441939874144313)
(0.4494036953479999,0.40883879748288582)
(0.50144522339199993,0.20441939874144269)
(0.55348675143600001,0.051104849685360887)
(0.60552827947999988,0.25552424842680443)
(0.65756980752399974,0.15331454905608202)
(0.70961133556799982,0.051104849685360672)
(0.76165286361199991,0.10220969937072177)
(0.81369439165599977,0.051104849685360568)
(0.86573591969999997,0)
};
\addlegendentry{Impostor (n=376)}
\addplot[ybar interval,area legend,fill=plotcolor2,fill opacity=0.5,draw=none] coordinates {
(-0.43530228139999999,0)
(-0.38326075335600002,0)
(-0.33121922531199999,0)
(-0.27917769726800001,0)
(-0.22713616922400001,0.40883879748288604)
(-0.17509464118000001,0.40883879748288626)
(-0.12305311313600004,0.40883879748288582)
(-0.071011585092000007,0.40883879748288626)
(-0.018970057048000033,0)
(0.033071470995999941,0.40883879748288582)
(0.08511299903999997,0.81767759496577253)
(0.13715452708399994,0.40883879748288626)
(0.18919605512799992,0.81767759496577253)
(0.24123758317199989,0.40883879748288537)
(0.29327911121599998,0.81767759496577253)
(0.34532063925999995,0.40883879748288626)
(0.39736216730399992,0.40883879748288626)
(0.4494036953479999,2.0441939874144293)
(0.50144522339199993,1.2265163924486562)
(0.55348675143600001,2.0441939874144355)
(0.60552827947999988,0)
(0.65756980752399974,0.40883879748288537)
(0.70961133556799982,0.81767759496577075)
(0.76165286361199991,3.2707103798630968)
(0.81369439165599977,3.2707103798630763)
(0.86573591969999997,0)
};
\addlegendentry{Genuine (n=47)}
\end{axis}
\end{tikzpicture}
\caption{Gate fusion scores for new masked ROF (seed 42). Common bins. Within-class normalisation. Genuine and impostor counts are shown.}
\label{fig:checked5}
\end{figure}
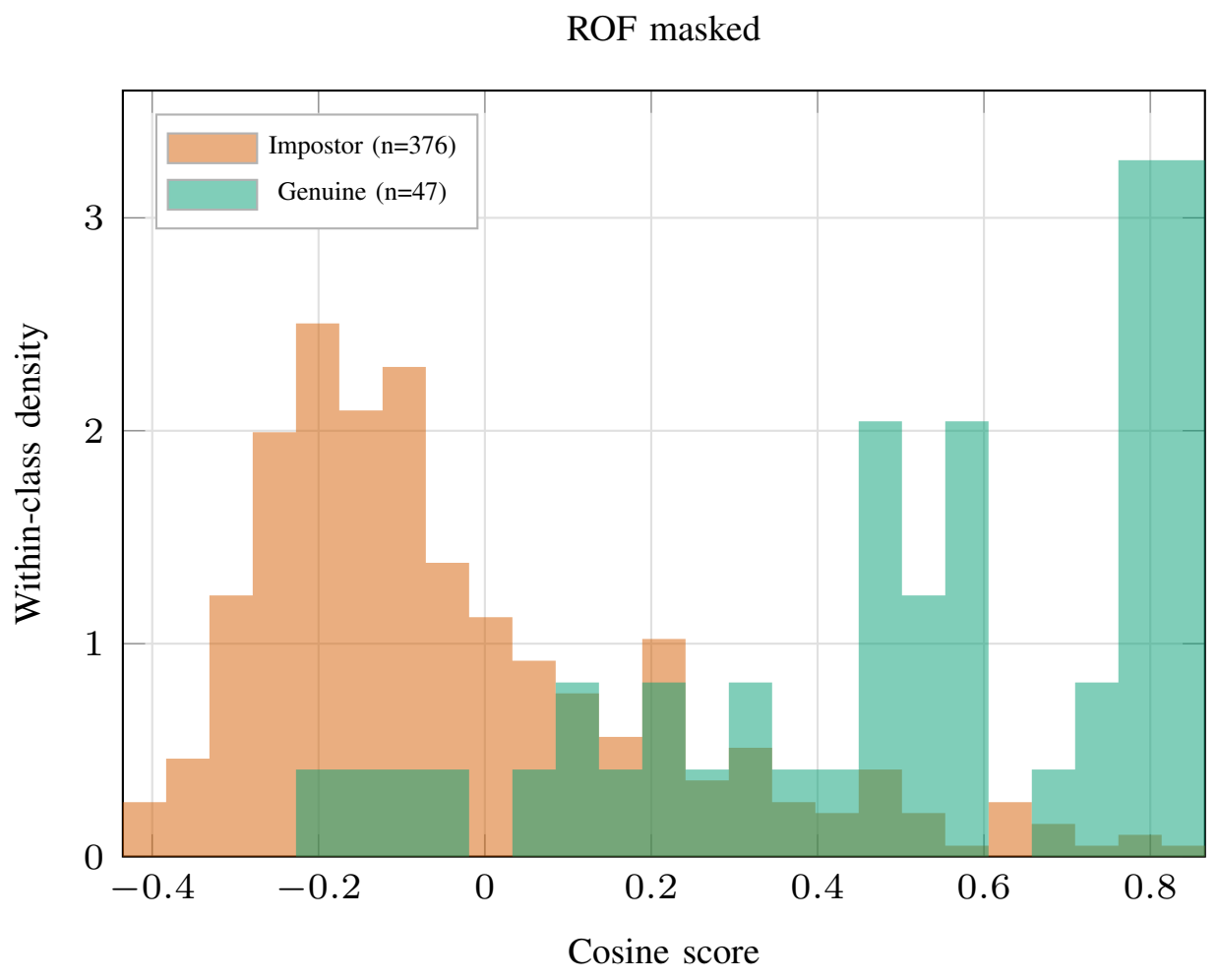

\subsection{Pretrained face-recognition diagnostic}
\textit{Aligned SFace diagnostic.} OpenCV SFace \cite{sface} (December 2021) was evaluated with YuNet \cite{yunet} (March 2023), five-landmark alignment, and 112-by-112-pixel native input. The detection confidence threshold was 0.8; the detection nearest the image centre was selected. No target fine-tuning was performed. Failed-pair scores were set to -1 and retained in the denominator; detected-only performance was subsequently analysed. The aligned protocol is detailed in Table~\ref{tab:checked11}. Masked, sunglasses and neutral ROF, plus retained MFR2 and RMFRD, achieved AUC values of 0.9399, 0.9440, 0.9991, 0.9968 and 0.8187. Corresponding failed-pair counts were 9, 48, 0, 0 and 1155. Endpoint information and raw scores are available within the released ledger.\par
Although these AUCs are larger than those of the evaluated DINOv2 models, they are not entirely comparable as pre-training, identity exposure, detection, alignment and optimisation goals differ. Hence, this ablation study does not establish state-of-the-art verification. In order to further facilitate model comparison, a controlled modern facial recognition baseline together with externally validated annotations would be advantageous.\par
\begin{table}[t]\centering
\caption{Aligned SFace with YuNet detection and five-landmark alignment was applied. Failed pairs were scored -1 and are accounted for in the denominator. Please note that these are not shared-crop SFace results (see Supplemental Table S3). There are differences in pre-training and pre-processing, so a controlled analysis of neural network architecture cannot be performed.}
\label{tab:checked11}
\footnotesize\setlength{\tabcolsep}{2pt}
\sbox{\uploadedtablebox}{\begin{tabular}{lll}\toprule
Protocol & AUC & Failed Pairs\\\midrule
Masked & 0.9399 & 9\\
Sunglasses & 0.9440 & 48\\
Neutral & 0.9991 & 0\\
MFR2 & 0.9968 & 0\\
RMFRD & 0.8187 & 1155\\
\bottomrule\end{tabular}}
\ifdim\wd\uploadedtablebox>\columnwidth\resizebox{\columnwidth}{!}{\usebox{\uploadedtablebox}}\else\usebox{\uploadedtablebox}\fi
\end{table}
\textit{Shared-crop SFace pre-processing.} The same pre-processed DINOv2 full-face input files were resized to generate 112-by-112-pixel SFace inputs. Neither YuNet detection nor five-landmark alignment was applied. AUCs were analysed in the same protocol sequence and were calculated as 0.8408, 0.7775, 0.8689, 0.9513 and 0.6532 (Supplemental Table S3). Embeddings were successfully generated for all 3402 unique images. The two SFace protocols involve different pre-processing, so their AUCs depend upon this and are not directly interchangeable. Neither can therefore be seen as a controlled training counterpart to DINOv2, since architecture, pre-training, resolution and supervision differ.\par
\subsection{Model pairings}
All pre-established pairings are detailed in Table~\ref{tab:checked12}. Negative deltas are also shown. Where a delta is positive, model 1 outperformed model 2. The same identity resampling weights were used for each model and all 3 seeds within each replicate. Mean delta across the seeds was then determined for each replicate. These intervals are dependent on the trained checkpoints and prescribed protocols and so are not performance confidence intervals for arbitrary populations.\par
\begin{table}[t]\centering
\caption{Mean delta AUCs. 95\% sensitivity intervals based on two-endpoint identity resampling. 1000 replicates. (G-F: Gate - Fixed fusion; W-B: Whole LoRA - Frozen base).}
\label{tab:checked12}
\footnotesize\setlength{\tabcolsep}{2pt}
\sbox{\uploadedtablebox}{\begin{tabular}{lllll}\toprule
Protocol & Compared & Delta & Lower & Upper\\\midrule
Masked & G-F & -0.0145 & -0.0560 & 0.0144\\
Masked & W-B & 0.0549 & -0.0424 & 0.1697\\
Sunglasses & G-F & -0.0026 & -0.0099 & 0.0061\\
Sunglasses & W-B & 0.0133 & -0.0030 & 0.0366\\
Neutral & G-F & 0.0030 & -0.0046 & 0.0106\\
Neutral & W-B & 0.0085 & -0.0117 & 0.0274\\
MFR2 & G-F & 0.0150 & 0.0044 & 0.0283\\
MFR2 & W-B & 0.0948 & 0.0513 & 0.1474\\
RMFRD & G-F & -0.0066 & -0.0135 & 0.0004\\
RMFRD & W-B & 0.0088 & -0.0067 & 0.0263\\
\bottomrule\end{tabular}}
\ifdim\wd\uploadedtablebox>\columnwidth\resizebox{\columnwidth}{!}{\usebox{\uploadedtablebox}}\else\usebox{\uploadedtablebox}\fi
\end{table}
Whilst subsets are analysed, positive and negative intervals fail to counter selection bias. Strict rules were set for data exclusion in these new runs. However, these related to problems identified in prior experimentation. Re-training was prospective under the cleaned protocol, but the exclusions were informed by earlier experiments. This does not confirm that local reconstructions are from an independent sample. With reference to provenance details, it is straightforward to recognise that the quantities in this experiment differ from those in the original full-case protocol.\par
\section{Discussion}
\subsection{Adaptation and regional complementarity}
In all test protocols, whole face LoRA has a higher mean AUC than frozen projection models, supporting hypothesis H1.\par
Depending on the area of the face shown, RQ2 is somewhat supported. Periocular LoRA has the best results in masked, but performs poorly in sunglasses and loses useful information in neutral.\par
RQ3 is less well supported. Gated models perform best in MFR2 and neutral ROF, but not in masks, sunglasses or RMFRD. Many of the comparisons included 0 within the range. Most importantly, the control experiments using the same branch show that the decision within gated models do not consistently form high quality representations. Using equal weighting in the models increased performance in masked compared to trained gated branches. Using a constant mean weighting, derived from validation only, also had competitive performance in ROF. There is no evidence to suggest that learning a gate automatically leads to increased robustness.\par
\subsection{AUC and calibration}
The AUC indicates the performance across the score range. However, in deployment a particular threshold will be selected. Higher AUC does not guarantee good performance at strict FAR requirements. The test counts at validation-calibrated thresholds indicate distributional shift. The thresholds that give desired performance in validation do not necessarily give desired performance on new identities. This is especially important in the neutral tests. Many identities were repeated in enrolment and probing, meaning that large pair counts appear to be present. However, this does not necessarily mean that large amounts of independent information is present.\par
Very few genuine pairs are present in the masked test (47). As only 9 identities are present, a change in 1 genuine acceptance will change TAR by 1/47. Also, a change in 1 impostor acceptance in testing will change FAR by 1/376. Four decimal places have been added to aid manual checks, but this does not indicate that four decimal place statistical precision has been reached. Identity node resampling is a sensitivity analysis for the limited protocol. This does not indicate that more independent identities and sessions have been obtained.\par
Score distribution plots show the overlapping of genuine and impostor scores.\par
\subsection{Positioning and negative findings}
Due to differences in training data and alignment, a fair comparison of model architectures could not be made.\par
Since a constant auxiliary vector is used, care must be taken in the interpretation of results.\par
\section{Limitations and Threats to Validity}
Despite removing duplicate pairs, it cannot be guaranteed that all labels are semantically correct. Also, not all perceptual duplicates may have been identified. Despite matching annotation sources, local differences in image capture may arise. Normalised name overlaps may fail to identify all alias pairs. Public figures may have been included in pre-training in DINOv2/SFace. It is unknown how many new identities they were exposed to.\par
Only 3 initialisations were tested. Seed SD provides descriptive information. Conditional sensitivity intervals from node bootstrapping depend on the 3 trained models and frozen identity graph. Identity sampling cannot add missing capture sessions or improve mislabelled images. With 10 comparisons, no control is provided for family wise error. Due to few masked identities and limited impostors, strong claims cannot be made in terms of security in cases of very low FAR.\par
RMFRD is exploratory. Although one known overlap has been removed in MFR2, it cannot be guaranteed that all identities in MFR2 were unseen. Differences in alignment and pre-training in SFace mean that model architectures could not be fairly compared. Maximum memory usage was not recorded. Training times (wall-clock) include pauses and sleeps and therefore should not be considered as indicative of continuous processing speeds. Deterministic training was performed on CPU. Checkpoint recovery was tested within the recorded run time. Byte for byte exact replication cannot be guaranteed between versions of packages, or between different hardware.\par
\section{Reproducibility}
Mathematical checks can be made without torch, images and checkpoints.\par
\noindent\textit{Execution and reproduction details: supplemental material.}
The release contains code, manifests, hashes, derived scores, and manuscript figures; source face images and trained checkpoint binaries are excluded.\par
\section{Conclusion}
Five parameter efficient configurations, both for whole faces and periocular regions, have been compared. None of the configurations provide the best performance across all conditions. Periocular LoRA had the best performance in masked (Mean AUC = 0.9221). Fixed fusion performed best in sunglasses (0.8950) and exploratory RMFRD (0.7240). Learned fusion had the best performance in neutral ROF (0.9458) and retained MFR2 (0.9548).\par
Care should be taken when interpreting these results. Removing one identity at a time preserves the masked gated-versus-fixed direction, but results depend on cohort composition. Results are also vulnerable to pre-processing as common crops were used in SFace. Despite useful individual features being learnt by the model, optimal weighting may not have been achieved. Constant weighting models may perform well. The variation in testing FAR values could be caused by threshold calibration during validation. It has been supported that complementary information exists in different regions, but it cannot be universally claimed that security and robustness is guaranteed.\par
Validation data is also required. Greater generality requires more distinct identities, per-face reference data and validated semantic information. Information on data origins together with associated ethics and permission details can be found in Supplemental Section VIII.\par
\section*{Acknowledgment}
The author prepared the manuscript and conducted all training and testing. OpenAI's ChatGPT provided coding assistance. The author takes responsibility for the code, results and final manuscript.\par
\section*{Funding and Competing Interests}
There was no external funding for this work. There are no competing interests to declare.\par

\clearpage
\twocolumn[{\centering\large\bfseries Supplemental Material\par\vspace{0.5em}}]
\renewcommand{\theHsection}{supp.\arabic{section}}
\renewcommand{\theHsubsection}{supp.\arabic{section}.\arabic{subsection}}
\renewcommand{\theHtable}{supp.\arabic{table}}
\renewcommand{\theHfigure}{supp.\arabic{figure}}
\setcounter{section}{0}
\setcounter{subsection}{0}
\setcounter{table}{0}
\setcounter{figure}{0}
\renewcommand{\thetable}{S\arabic{table}}
\renewcommand{\thefigure}{S\arabic{figure}}

\begin{table}[t]\centering
\caption{THE TOTAL NUMBER OF PARAMETERS WITHIN MODELS AND PROPORTION TRAINABLE IS GIVEN IN SUPPLEMENTAL TABLE S1. TRAINING ONLY CLASSIFIER HEAD NOT INCLUDED.}
\label{tab:checked2}
\footnotesize\setlength{\tabcolsep}{2pt}
\sbox{\uploadedtablebox}{\begin{tabular}{llll}\toprule
ID & Total Parameters & Trainable Parameters & (\%) Trainable\\\midrule
M0 & 87171072 & 590592 & 0.68\\
M1 & 87613440 & 1032960 & 1.18\\
M2 & 175226880 & 2065920 & 1.18\\
M3 & 175226880 & 2065920 & 1.18\\
M4 & 175424257 & 2263297 & 1.29\\
\bottomrule\end{tabular}}
\ifdim\wd\uploadedtablebox>\columnwidth\resizebox{\columnwidth}{!}{\usebox{\uploadedtablebox}}\else\usebox{\uploadedtablebox}\fi
\end{table}

\begin{table}[t]\centering
\caption{AVERAGE WEIGHTING IN FACE BRANCH FOR EACH PROTOCOL AND SEED. STANDARD DEVIATION (IN BRACKETS) ACROSS UNIQUE IMAGE ENDPOINTS. WEIGHTS ARE DESCRIPTIVE AND DO NOT INDICATE CALIBRATED RELIABILITY.}
\label{tab:checked10}
\footnotesize\setlength{\tabcolsep}{2pt}
\sbox{\uploadedtablebox}{\begin{tabular}{llll}\toprule
Protocol & Seed 7 & Seed 42 & Seed 123\\\midrule
Masked & 0.712 (0.118) & 0.709 (0.100) & 0.749 (0.092)\\
Sunglasses & 0.841 (0.082) & 0.825 (0.083) & 0.830 (0.082)\\
Neutral & 0.773 (0.107) & 0.767 (0.096) & 0.778 (0.099)\\
MFR2 & 0.659 (0.118) & 0.632 (0.117) & 0.696 (0.103)\\
RMFRD & 0.745 (0.086) & 0.697 (0.090) & 0.775 (0.071)\\
\bottomrule\end{tabular}}
\ifdim\wd\uploadedtablebox>\columnwidth\resizebox{\columnwidth}{!}{\usebox{\uploadedtablebox}}\else\usebox{\uploadedtablebox}\fi
\end{table}

\noindent\textit{References: main-paper numbering.}\par

\section{Parameters}

M0 has 590,592 trainable parameters. M2 includes unused full face modality. There are 1,032,960 trainable parameters in the active periocular branch. The nominal number of trainable parameters in this model instance is 2,065,920. The 128 class head used during training only has 98,432 parameters. The maximum number of trainable parameters across all adapted models is 2.26M (1.29\%). This value occurs in a double encoder model (175M). Hardware details, runtime durations, software versions, and hash values of checkpoints are stored in the artifact. Maximum memory usage is not recorded.\par

\section{Run and Recovery}

This was run in PyTorch (v2.14.0+cpu). The weights were cached from DINOv2. Atomic checkpoints were used containing optimizer states, ordered data samples, index of next batch, loss in current epoch, and random number generator states (Python, NumPy, PyTorch). Checkpoints which are incompatible with the current run (by mismatched manifest and source signature) cannot be recovered. In order to retain an evaluated run as complete in the serial co-ordinator, all required hash comparisons must pass. Periocular only training means that the unused full face forward pass was skipped during training. As such random numbers would have been sampled differently to the historical trainer. It cannot therefore be stated that bitwise identical optimization was carried out.\par

\section{Reproducibility}

Data with licences must be used to reproduce this work. Face / periocular image paths must be identical. Training and pair manifests must be frozen. Software must be set up as described, and cached DINOv2 weights loaded. Models will train for 5 epochs. Models will be scored using each protocol. During the sweep partial model summaries may be created, but all models must be assessed to produce final pair comparisons. File locking has been implemented to ensure only one instance of the coordinator is run in the recorded Windows environment. If started multiple times, verified complete runs will be retained, or incomplete runs will be continued. If stopped, runs will continue from the latest saved state of the model, including optimiser and random number generator states. Unsaved work may be re-run. Face images, pre-trained binary weights and newly trained checkpoints (\textasciitilde{}8.6GB) are not included. However, checksums to verify provenance within your local environment are included. Binary files must also be provided to reproduce inference.\par

Mathematical validation is possible without torch, images and checkpoints. ROC metrics can be recalculated from raw scores within this script, and it can be confirmed that pairs are correct and checksums match. Mean and standard deviation for seed values can be recalculated, and contents of the table in the manuscript can be validated. 120 raw score files from completed CPU runs, sensitivity files and SFace controls are included. If all of these validate, then internal consistency of the validated files can be confirmed. However, there is no guarantee of novelty, acceptance for publication, or semantic correctness of image labelling within this work.\par

Earlier versions of this manuscript and subsequent work can be found in the Git history. Some claims and numerical data within these files are now out of date when compared with the completed CPU experiments in this paper. Newly trained models are not represented in the filtered scores within previous files. Incomplete training / smoke tests will not produce headline tables. Tables and vector plots within IEEE and arXiv versions of this work have been made consistent with recorded results.\par

\section{Further Analysis \& Results}

For all testing protocols, the full face LoRA models have higher mean AUCs than frozen projection models, meaning H1 is supported for mean AUC. Largest performance improvements were made in MFR2 and masked ROF. Smaller improvements were observed in the RMFRD. Supervised adaptation seems to be useful in this experiment. It cannot be determined whether generalized visual encoder adaptation (rather than adaptation of face specific pre-trained models) is useful in this experiment.\par

Depending on the area of the face shown, the answer to RQ2 varies. Periocular LoRA has the highest mean AUC in masked ROF, but performs poorly in sunglasses and loses useful information in neutral. Therefore H2 is only somewhat supported. Upper face features provide useful information for distinguishing between subjects in lower face occlusion scenarios. However, they do not provide a good representation in all 3 conditions. Using fixed fusion provides a good baseline since information in both regions can be preserved without learning a weighting policy.\par

From results in SFace, the aligned diagnostic has higher AUCs in these protocols. However, due to differences in training data and alignment, a fair comparison of model architectures could not be made. Therefore this work should be considered as an ablation study and study of reproducibility. In order to claim competition within face recognition, comparisons must be made with equivalent baseline models, with more identities and verified labels.\par

Since a constant auxiliary vector is used, care must be taken in the interpretation of results. As constant input descriptors are provided, no image-specific quality variation is supplied by them in this experiment. Although the gate can react to embeddings, no evidence can be provided that quality cues improve performance. Future work using quality cues should be designed with validated descriptors and controlled ablation studies on descriptors. Evaluation criteria should also be fixed before investigation.\par

Records for 15 models in the grid have been included. These include 120 ordered raw-score files across 8 validation/test/external protocols. There are also evaluation reports, final training progression information, checksums of checkpoints and a table summarising 5 conditions with 25 rows. There are 10 comparisons of AUC in pairs with 3 seeds. Variation in AUC between seeds and conditional identity-resampling sensitivity intervals is included. Fixed versions of trainer, evaluator, common utilities, models and metrics have been included. Checksums are provided for each. If any of these sources change, execution signatures will become invalid.\par

\section{SAME-BRANCH COMPARISON CONTEXT}

As different weights are applied in the fixed and gated fusion models, different features may be learnt in each branch. By applying different weights to the gated models, but keeping the same branches, it was found that equal weighting in the gated branches gives a higher AUC than learned weighting in masked ROF pairs. This analysis allows understanding of the representations and the inference weighting rule. However, no assumptions are made on whether the gate helps during joint training.\par

\section{EXTENDED INTERPRETATION AND SOURCE-IMAGE REVIEW}

The methods used to produce fusion, LoRA and weighted models are well known.\par

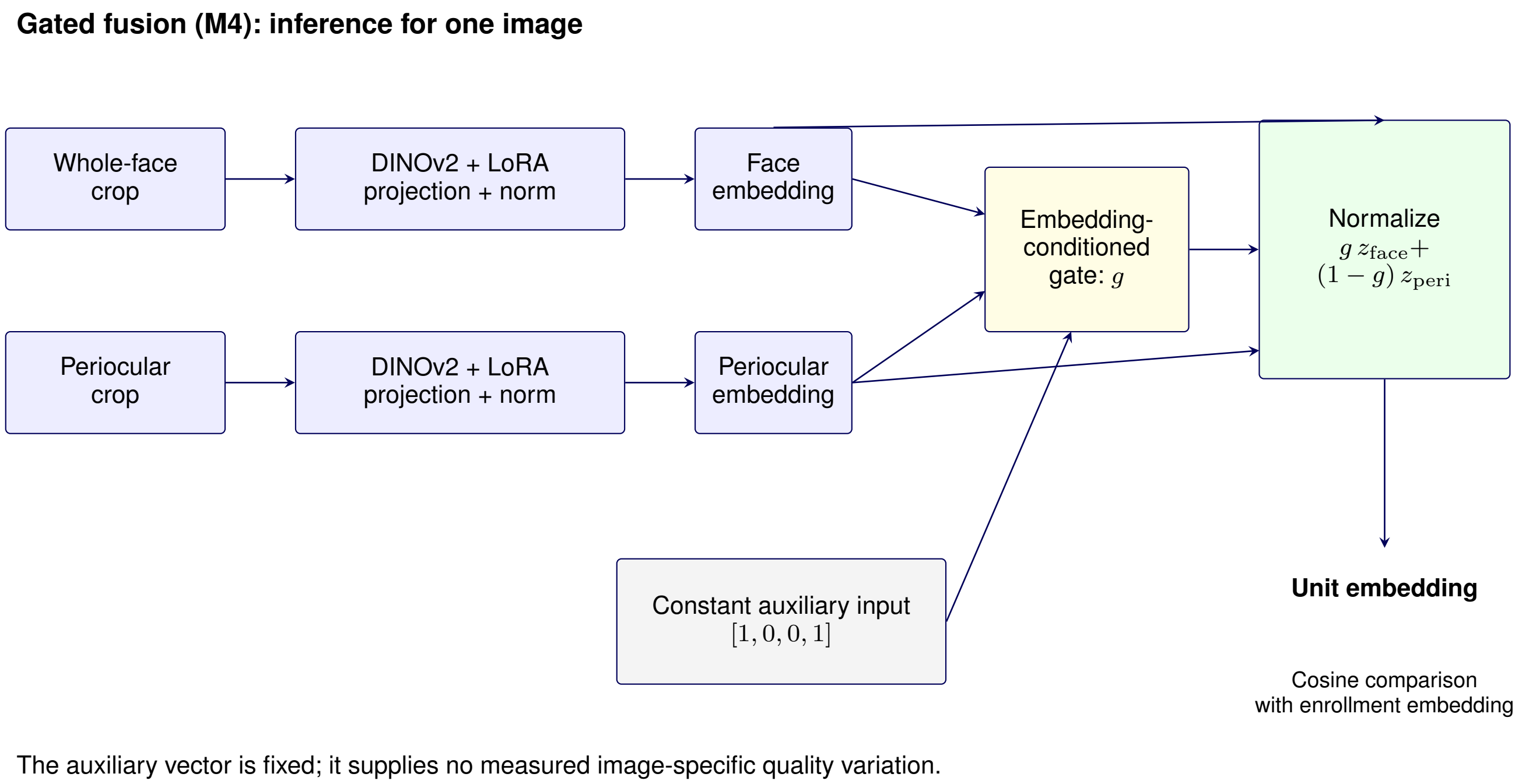
\begin{figure*}[t]\centering
\begin{tikzpicture}[x=0.098\textwidth,y=0.098\textwidth,
  every node/.style={font=\sffamily\footnotesize,align=center},
  block/.style={draw=blue!35!black,rounded corners=1.5pt,fill=blue!7,inner sep=2pt},
  flow/.style={->,>=stealth,draw=blue!35!black,line width=0.5pt}]
\path[use as bounding box] (0,0) rectangle (10,5.2);
\node[anchor=west,font=\sffamily\small\bfseries] at (0.15,4.85)
  {Gated fusion (M4): inference for one image};
\node[block,minimum width=0.1372\textwidth,minimum height=0.0637\textwidth] (face) at (0.85,3.875) {Whole-face\\crop};
\node[block,minimum width=0.1372\textwidth,minimum height=0.0637\textwidth] (peri) at (0.85,2.575) {Periocular\\crop};
\node[block,minimum width=0.2058\textwidth,minimum height=0.0637\textwidth] (ef) at (3.05,3.875) {DINOv2 + LoRA\\projection + norm};
\node[block,minimum width=0.2058\textwidth,minimum height=0.0637\textwidth] (ep) at (3.05,2.575) {DINOv2 + LoRA\\projection + norm};
\node[block,minimum width=0.098\textwidth,minimum height=0.0637\textwidth] (zf) at (5.05,3.875) {Face\\embedding};
\node[block,minimum width=0.098\textwidth,minimum height=0.0637\textwidth] (zp) at (5.05,2.575) {Periocular\\embedding};
\node[block,fill=gray!9,minimum width=0.2058\textwidth,minimum height=0.0784\textwidth] (q) at (5.10,1.05) {Constant auxiliary input\\$[1,0,0,1]$};
\node[block,fill=yellow!12,minimum width=0.1274\textwidth,minimum height=0.1029\textwidth] (gate) at (7.05,3.425) {Embedding-\\conditioned\\gate: $g$};
\node[block,fill=green!8,minimum width=0.1568\textwidth,minimum height=0.1617\textwidth] (norm) at (8.95,3.425) {Normalize\\$g\,z_{\mathrm{face}}+$\\$(1-g)\,z_{\mathrm{peri}}$};
\draw[flow] (face.east) -- (ef.west);
\draw[flow] (peri.east) -- (ep.west);
\draw[flow] (ef.east) -- (zf.west);
\draw[flow] (ep.east) -- (zp.west);
\draw[flow] (zf.east) -- (6.4,3.65);
\draw[flow] (zp.east) -- (6.4,3.16);
\draw[flow] (q.east) -- (6.95,2.9);
\draw[flow] (gate.east) -- (norm.west);
\draw[flow] (zf.north) -- (8.95,4.25);
\draw[flow] (zp.east) -- (8.15,2.78);
\draw[flow] (norm.south) -- (8.95,1.52);
\node[font=\sffamily\footnotesize\bfseries] at (8.95,1.25) {Unit embedding};
\node[anchor=north,font=\sffamily\scriptsize] at (8.95,0.8)
  {Cosine comparison\\with enrollment embedding};
\node[anchor=west,font=\sffamily\footnotesize] at (0.15,0.12)
  {The auxiliary vector is fixed; it supplies no measured image-specific quality variation.};
\end{tikzpicture}
\caption{In this case, there is no variability in the auxiliary quality descriptors. Fig. S1 shows the inference flowchart for this model. The embeddings from both branches, along with the constant auxiliary vector, are passed into the gate, which weights the two branch embeddings. The weighted sum is normalised. In the enrolment image, the model is identical. During inference, the training-only identity classifier has been removed.}\label{fig:supparchitecture}\end{figure*}

Having explored DINOv2-LoRA periocular models [25] and face recognition models [30], it can no longer be stated that novel contributions to this work come from adaptation within the backbone. However, it remains interesting to investigate whether a similar robustness can be achieved with regionally extracted features, weights and thresholds. When separately trained models are compared, differences cannot be attributed to inference weights alone. Having identical control models in each branch is a better indicator of this.\par

Score distributions allow observation of overlap between genuine and impostor scores. However, the plots provide no information as to which identities caused these errors, or why they occurred. Qualitative conclusions must be made by manual inspection of annotated identities in licensed source data. This is possible from the manifests and exclusion ledgers saved, meaning that private/ restricted facial images do not need to be distributed in the repository. Since a constant auxiliary vector is used, caution should be taken in interpreting results. Since constant input descriptors have been provided, no measured image-specific quality variation is supplied by them in this experiment. While the gate can be influenced by embeddings, it cannot be proved that quality information improves performance.\par

\section{Common Preprocessing for SFace}

The exact DINOv2 full face files were scaled to 112x112 to create SFace images, with no YuNet detection or landmark alignment. AUC values of 0.8408, 0.7775, 0.8689, 0.9513 and 0.6532 were found (in this order) (See Supplemental Table S3). Features were successfully extracted from all 3402 unique images. Due to differing preprocessing between protocols, the AUC values cannot be considered directly comparable baseline values. Neither protocol is a fair comparison to DINOv2 at training due to differences in architecture, pre-training, resolution and supervision.\par

\begin{table}[!ht]\centering\footnotesize\setlength{\tabcolsep}{2pt}
\caption{AUCs from different preprocessing protocols applied to SFace. All pairs with failed comparisons were included in AUC (score -1). Number of failed pairs per protocol is shown. Only aligned comparison diagnostic results are shown in Table IX within the main text.}
\label{tab:sfaceshared}\begin{tabular}{lrrrr}\toprule
Protocol & Aligned AUC & Failed & Shared AUC & Failed\\\midrule
Masked ROF & 0.9399 & 9 & 0.8408 & 0\\
Sunglasses ROF & 0.9440 & 48 & 0.7775 & 0\\
Neutral ROF & 0.9991 & 0 & 0.8689 & 0\\
MFR2 & 0.9968 & 0 & 0.9513 & 0\\
RMFRD & 0.8187 & 1155 & 0.6532 & 0\\
\bottomrule\end{tabular}\end{table}

\section{Provenance}

Data from existing sources have been used within this project (ROF, MFR2, RMFRD). The Kaggle execution copies can be identified. The source data can be found in the ROF and RMFRD repositories and the MFR2 maintainer page. Within the local ROF annotations, the source audited files are the same. A complete trace as to how the images were obtained was not able to be found. Identifiers relating to source data and exclusion ledgers remain.\par

Within the ROF repository, software and documents are licensed under MIT. In the MFR2 maintainer page, Creative Commons licensing is mentioned but no further details on the type or terms of the licence are given. In the RMFRD README, it is stated that the data was obtained from online sources. No clear licence information was found, however, regarding using the images. It cannot be guaranteed that all images used have appropriate licenses, and redistribution rules have been applied appropriately. Source data for faces, trained model checkpoints were not provided in this release. Source code, manifests, hashes, processed scores and figures from the manuscript have been provided. This project was not assessed by an ethical review board. There was no formal approval or exemption obtained. Consent from individuals or full permission for images in this work could not be verified from the data reviewed. From the information publicly available, no approvals, waivers, consent or general permission to use images in this data can be inferred. It cannot be ascertained if ethical considerations were made, and permission to use certain datasets/images was given.\par

\end{document}